\documentclass{article}
\usepackage{wrapfig}
\usepackage[preprint]{neurips_2025}
\usepackage[utf8]{inputenc}
\usepackage[T1]{fontenc}    
\usepackage{hyperref}       
\usepackage{url}            
\usepackage{amsmath}
\usepackage{booktabs}       
\usepackage{subcaption}
\usepackage{amsfonts}
\usepackage{amsfonts}       
\usepackage{nicefrac}       
\usepackage{amssymb}  
\usepackage{bm} 
\usepackage{microtype}      
\usepackage{xcolor}         
\usepackage{multirow} 
\usepackage{amsmath, amssymb}
\usepackage{amsthm}
\usepackage{picinpar}
\usepackage{footmisc}
\usepackage{amsfonts}
\usepackage{graphicx}
\usepackage{amssymb}
\usepackage{colortbl}
\usepackage{sidecap}
\usepackage{array}
\newcommand{\appendixtitle}[1]{
    \begin{center}
    \hrule height 3.5pt  
    \vspace{1.2em}
    \textbf{\LARGE #1}  
    \vspace{1.4em}
    \hrule height 1pt  
    \vspace{1.5em}
    \end{center}
}
\title{Examining the Source of Defects from a Mechanical Perspective for 3D Anomaly Detection}

\author{%
  Hanzhe Liang\\
  Shenzhen University
  \And
  Aoran Wang\\
  Shanghai AI Lab
  \And
  Jie Zhou \\
  Shenzhen University
  \And
  Xin Jin \\
  Ningbo EIT
  \And
  Can Gao$^{\dag}$ \\
  Shenzhen University
  \And
  Jinbao Wang$^{\dag}$\\
  Shenzhen University
}

\begin{document}

\maketitle
\renewcommand{\thefootnote}{\fnsymbol{footnote}}
        \footnotetext{$\dag$ Co-corresponding author. \textit{This paper and code were done primarily by Hanzhe Liang while at Shenzhen University. For questions, please send an email to} \url{2023362051@email.szu.edu.cn}.}

\begin{figure}[!ht]
    \centering
    \includegraphics[width=0.99\linewidth]{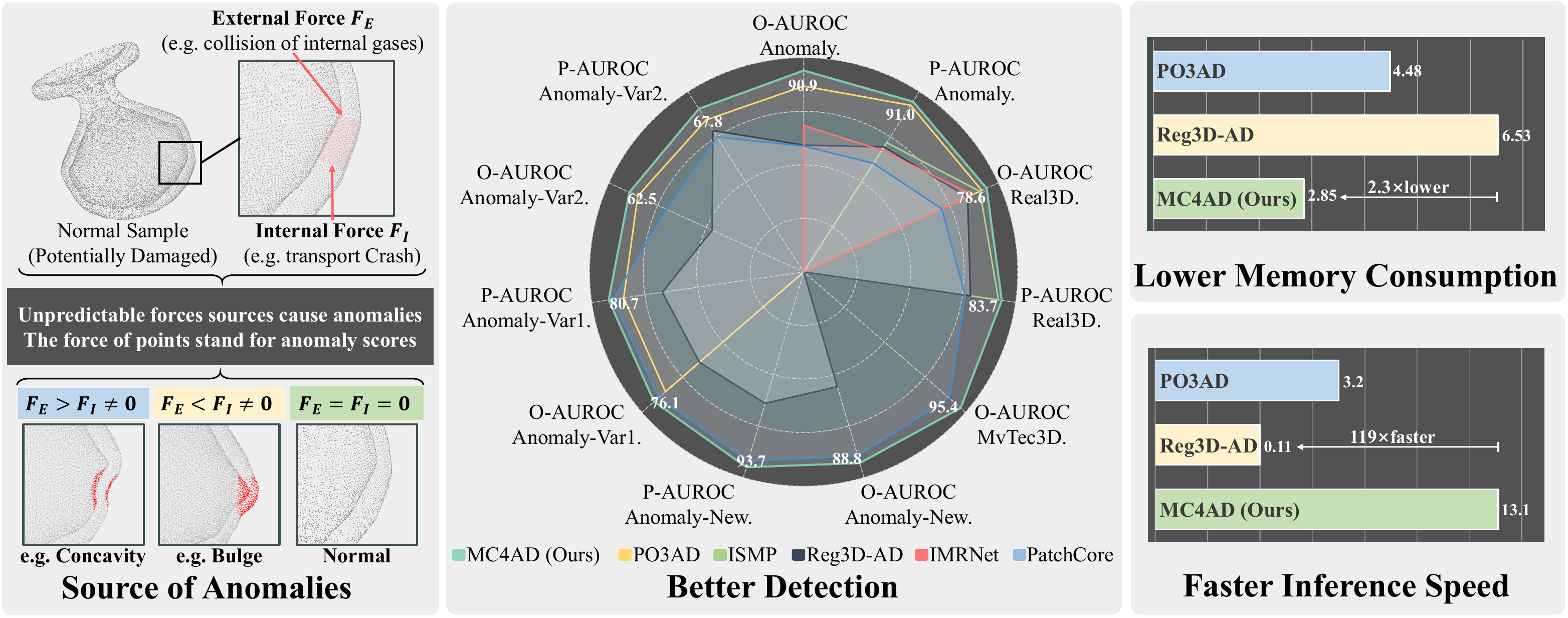}
    \caption{\textbf{Overview of our MC4AD.} Compared with existing methods, the proposed MC4AD demonstrates excellent performance, low memory, and more efficiency. We examine for the first time the source of anomalies from a mechanical perspective and use it as a motivation to design models.}
    \label{abs}
\end{figure}

\begin{abstract}

In this paper, we explore a novel approach to 3D anomaly detection (AD) that goes beyond merely identifying anomalies based on structural characteristics. Our primary perspective is that most anomalies arise from unpredictable defective forces originating from both internal and external sources. To address these anomalies, we seek out opposing forces that can help correct them. 
Therefore, we introduce the Mechanics Complementary Model-based Framework for the 3D-AD task (MC4AD), which generates internal and external corrective forces for each point. We first propose a Diverse Anomaly-Generation (DA-Gen) module designed to simulate various types of anomalies. Next, we present the Corrective Force Prediction Network (CFP-Net), which uses complementary representations for point-level analysis to simulate the different contributions from internal and external corrective forces.
To ensure the corrective forces are constrained effectively, we have developed a combined loss function that includes a new symmetric loss and an overall loss. Notably, we implement a Hierarchical Quality Control (HQC) strategy based on a three-way decision process and contribute a dataset titled Anomaly-IntraVariance, which incorporates intraclass variance to evaluate our model.
As a result, the proposed MC4AD has been proven effective through theory and experimentation.
The experimental results demonstrate that our approach yields nine state-of-the-art performances, achieving optimal results with minimal parameters and the fastest inference speed across five existing datasets, in addition to the proposed Anomaly-IntraVariance dataset.
The source is available at this \href{https://github.com/hzzzzzhappy/MC4AD}{link}.

\end{abstract}

\section{Introductuon}
The task of 3D anomaly detection (3D-AD) in industrial manufacturing requires identifying anomalies within 3D point clouds, representing structural deviations from the norm~\cite{LIN2025103139,Li_2024_CVPR,NEURIPS2023_611b896d,2025arXiv250205779J}. 
Existing methods are mainly classified into memory bank methods~\cite{liu2023real3d,ismp,Zhu_2024} and reconstruction methods~\cite{r3dad,Li_2024_CVPR,liao2025robustdistributionalignmentindustrial}. Memory bank methods store the normal structural features into the memory bank during the training phase and judge the anomalies by comparing the structural features to be tested with the memory bank features during the testing phase~\cite{BTF,ismp,Zhu_2024,liu2023real3d}. Reconstruction methods train the model to eliminate the abnormal structures and detect the anomalies by comparing the original and reconstructed structures~\cite{Li_2024_CVPR,r3dad,PO3AD,cao2025}.

\textbf{Motivation.} 
Deformation defects such as bulges, cracks, and scratches often occur on product surfaces during industrial production. These defects can be attributed to unexpected mechanical actions during production and transportation. \textit{While existing techniques mostly focus on detecting structural features, they do not consider the underlying causes of these anomalies.} To address this issue, this paper designs a novel Mechanics Complementary Model~(MCM) to simulate the interaction of these mechanical forces, as shown in Section~\ref{sec:mcd}. The proposed framework~(MC4AD) employs the MCM for the 3D-AD task, which specifically considers anomalies resulting from mechanical forces acting during production. These forces, including internal and external defective forces, are modeled and counteracted by corrective forces. This process aims to restore the point cloud to its normal form, as illustrated in Figure~\ref{abs}.

In MC4AD, we propose the Diverse Anomaly-Generation (DA-Gen) module, which simulates a variety of anomalies by perturbing the surface normal vectors of 3D objects. This approach allows for more realistic defect generation. In addition, we introduce the Corrective Force Prediction Network (CFP-Net), which employs complementary representations to predict corrective forces at the point level. Besides, we apply a combined loss function that includes both symmetric and reconstruction loss functions, aiming to restore the affected regions of the point cloud. The efficiency of our model is further demonstrated through Hierarchical Quality Control (HQC) strategies, which are designed to simulate real-world, cost-limited industrial environments. Extensive experiments conducted on multiple datasets show that our approach, MC4AD, achieves state-of-the-art (SOTA) performance in detection and segmentation, as illustrated in Figure~\ref{abs}.

Overall, the main contributions of this paper are summarized as follows:
\begin{itemize}
    \item To the best of our knowledge, we are the first to examine the causes of structural anomalies from a mechanical perspective. We consider it the result of internal and external defective forces, and we aim to correct them by simulating the opposing force.
    
    \item We propose a novel framework (MC4AD) that creates diverse pseudo anomalies with DA-Gen and introduces a CFP-Net that captures complementary features to predict internal and external corrective forces. A new combination of losses is proposed to optimize symmetry between the internal and external corrective forces.

    \item We propose an HQC strategy to simulate multiple detections on real assembly lines and construct a new dataset named Anomaly-IntraVariance to evaluate performance under cost-limited conditions, making it more relevant to real-world industry scenarios.

    \item Extensive experiments demonstrate the MC4AD's effectiveness. We achieved nine state-of-the-art results with minimal parameters and the fastest inference speed in detection and segmentation tests across five existing and proposed datasets involving 90 classes. 
\end{itemize}

\section{Related Work}
\textbf{2D Anomaly Detection.}
The existing 2D approaches mainly follow two paradigms: Generative and discriminative methods~\cite{Liu_2024,9849507,10443076}. Generative methods that model normal distributions using networks like Generative Adversarial Networks~\cite{song2021anoseganomalysegmentationnetwork,liang2022omni}, Autoencoders~\cite{9681338,Liu_2025} or Diffusion models~\cite{9857019,he2024diffusion} to identify deviations, and discriminative methods that leverage supervised learning to train defect classifiers~\cite{9879738,spade,efficientad,Bergmann_2020_CVPR,RudWeh2022}. There are morphological differences between images and point clouds, and applying 2D-AD techniques directly to 3D presents challenges.

\textbf{3D Anomaly Detection.}
The 3D-AD methods aim to identify and localize anomalous points or regions within 3D point cloud data~\cite{AST,shape,looking3d,liang2025fencetheoremdualobjectivesemanticstructure}. Existing methodologies are broadly categorized into feature embedding-based and Reconstruction-based Methods. Feature embedding-based methods extract discriminative features from normal samples and create normal distributions. The features to be tested are used to determine anomalies by their distance from the normal distribution~\cite{BTF,Zhu_2024,m3dm,liu2023real3d,ismp}.
Reconstruction-based methods compress point clouds into latent representations and reconstruct their original feature~\cite{liu2024splatposerealtimeimagebasedposeagnostic,Li_2024_CVPR,r3dad,PO3AD,Kruse_2024_CVPR}.
The large language model~(LLM) also demonstrates anomaly detection capabilities on the zero-shot task~\cite{2023arXiv231102782C, Wang_2025_WACV,pointad,AdaCLIP,ENEL,cheng2024zeroshotpointcloudanomaly}.
However, existing methods detect anomalies only from the morphological structure and never consider the source of the anomaly. This paper rethinks the source of anomalies from a mechanical perspective.

\section{Preliminaries}
\label{motivation}

\textbf{Definition A1.} Given a normal sample as 3-manifold $\texttt{M}\in \mathbb{R}^3$, which can be discretized as point cloud $P\text{=}\{\texttt{p}_{i=1}^n\}\in\mathbb{R}^{n\times3}$ with each point $\texttt{p}_i=(x_i,y_i,z_i)^\top$. 

\textbf{Definition A2.} Given a resultant defective force $F_D(\texttt{p})\in \mathbb{R}^3$, damaging normal object $\texttt{M}$ to anomalous object $\texttt{M}'$, which can be decompose as a resultant external defective force $F_E(\texttt{p})\in \mathbb{R}^3$ and a resultant internal defective force $F_I\in \mathbb{R}^3$. These forces can be defined as 
\begin{equation}
\begin{aligned}
    F_D(\texttt{p}) =\int_M f_D(\texttt{p}) \, d\texttt{M}
        =F_E(\texttt{p}) + F_I(\texttt{p})
        =\int_{S^{+}} f_E(\texttt{p}) \, dS^+ + \int_{S^{-}} f_I(\texttt{p}) \, d{S^{-}},
\end{aligned}
\end{equation}
where $f_D(\texttt{p})$, $f_E(\texttt{p})$, and $f_I(\texttt{p})$ represent the resultant defective, component external defective, and component internal defective force, respectively. $S^+$ and $S^{-}$ stand for external and internal planes.

\textbf{Definition A3.} Given a resultant corrective force $F_C\in \mathbb{R}^3$, restoring anomalous object $\texttt{M}'$ to normal object $\texttt{M}$, 
which can be decompose as a resultant external corrective force $F_E'\in \mathbb{R}^3$ and a resultant internal corrective force $F_I'\in \mathbb{R}^3$. These forces can be defined as
\begin{equation}
\begin{aligned}
    F_C(\texttt{p}) =\int_M f_C(\texttt{p}) \, d\texttt{M},
        =F_E'(\texttt{p}) + F_I'(\texttt{p})&=\int_A f_E'(\texttt{p}) \, dS^+ + \int_{S^{-}} f_I'(\texttt{p}) \, d{S^{-}}, \\
        =-F_E(\texttt{p})-F_I(\texttt{p})=-F_D(\texttt{p})&=\int_A -f_E(\texttt{p}) \, dS^+ + \int_{S^{-}} -f_I(\texttt{p}) \, d{S^{-}},\\
\end{aligned}
\end{equation}
where $f_D'(\texttt{p})$, $f_E'(\texttt{p})$, and $f_I'(\texttt{p})$ represent the component force, component external Corrective force, and component internal Corrective force, respectively. 

\textbf{Definition A4.} Given a non-linear additive map $\phi:(F,G)\rightarrow\nabla G$ to represent the affect of the focre $F$ to object $G$, which can be defined as:
\begin{equation}
\begin{aligned}
\texttt{M}+\phi(F_D(\texttt{p}),\texttt{M})
&=\texttt{M}+\phi(\texttt{M},\int_M f_D(\texttt{p}) \, d\texttt{M})=\texttt{M}+\phi(\texttt{M},\int_{S^{+}} f_E(\texttt{p}) \, dS^++\int_{S^{-}} f_I(\texttt{p}) \, dS^{-}),\\
&=\texttt{M}+\nabla\texttt{M}_E+\nabla\texttt{M}_I=\texttt{M}+\nabla\texttt{M}=\texttt{M}',\\
\texttt{M}'+\phi(F_C(\texttt{p}),\texttt{M}')
&=\texttt{M}+\phi(\texttt{M}',\int_{M'} f_D(\texttt{p}) \, d\texttt{M}')=\texttt{M}'+\phi(\texttt{M}',\int_{S^{+}} f_E'(\texttt{p}) \, dS^++\int_{S^{-}} f_I'(\texttt{p}) \, dS^{-}),\\
&=\texttt{M}+\nabla\texttt{M}_E'+\nabla\texttt{M}_I'=\texttt{M}'+\nabla\texttt{M}'=\texttt{M}.\\
\end{aligned}
\end{equation}
where $\nabla\texttt{M}=-\nabla\texttt{M}'$, meaning: \textit{any object damaged by a defective force $F_D$ can be restored by the opposite resultant Corrective force $F_C$, idealy}. We show the symbols used in Appenix~\ref{ms}.


\textbf{Mechanics Complementary Model.}\label{sec:mcd}
Mechanical actions include the \underline{external force} $F_{E}(\texttt{p})$ and \underline{internal force} $F_{I}(\texttt{p})$ mentioned in \textbf{Def. A2}, represented by collisions and gas expansion, respectively.
The composition of defective forces $F_{D}(\texttt{p})$ of external and internal forces causes defects mentioned in \textbf{Def. A2}. Anomalies arise from an imbalance in one of the internal or external forces. 
By simulating a composition Corrective force $F_C(\texttt{p})$ opposite to the $F_{D}$ mentioned in \textbf{Def. A3}, applying $F_{D}(\texttt{p})$ to the defect, the defect will ideally be restored as mentioned in \textbf{Def. A4}.
Based on the mechanical mechanism of defect formation, the conversion of the defective form $\texttt{M}'$ to the standard form $\texttt{M}$ is realized by applying the Corrective force $F_C(\texttt{p})$, which is inverse to the defective force $F_D(\texttt{p})$. We treat each point of the point cloud as a mechanical node subject to the defect force $F_D$: the theoretical force vector of a normal node should be a zero vector, while an abnormal node presents a non-zero vector. By predicting the modified force vector $F_C(\texttt{p})$ (which is synthesized by the corresponding external force component $F_E'(\texttt{p})$ and internal force component $F_I'(\texttt{p})$) at each node, the mechanical influence of $F_D(\texttt{p})$ can be reversed to cancel out, thus restoring the normal geometric characteristics of the abnormal node. The motivation is formalized as:
\begin{equation}\label{eq}
\begin{aligned}
\texttt{Damage Source}:F_D(\texttt{p}) &= \int_{S^{+}} f_E(\texttt{p}) , d{S^{+}} + \int_{S^{-}} f_I(\texttt{p}) , dS^{-}, \\
\texttt{Corrective Object}: F_C(\texttt{p}) &= \int_{S^{+}} f_E'(\texttt{p}) , d{S^{+}} + \int_{S^{-}} f_I'(\texttt{p}) , dS^{-} ,\\
\texttt{if} \ \texttt{p}\ \texttt{is normal} F_D(\texttt{p})&=\textbf{0}\ \texttt{else} \neq \textbf{0},\ F_C(\texttt{p})=-F_D(\texttt{p}).
\end{aligned}
\end{equation}
where \texttt{Damage Source} and \texttt{Corrective Object} represent the potential damage at each point, and the goal of the model is to restore the damage, respectively.
We further describe the theoretical implications and explain the connection to 3D anomaly detection in Appendix~\ref{moremoti}.

\section{Method}
\begin{figure}[ht]
    \centering
    \includegraphics[width=0.75\linewidth]{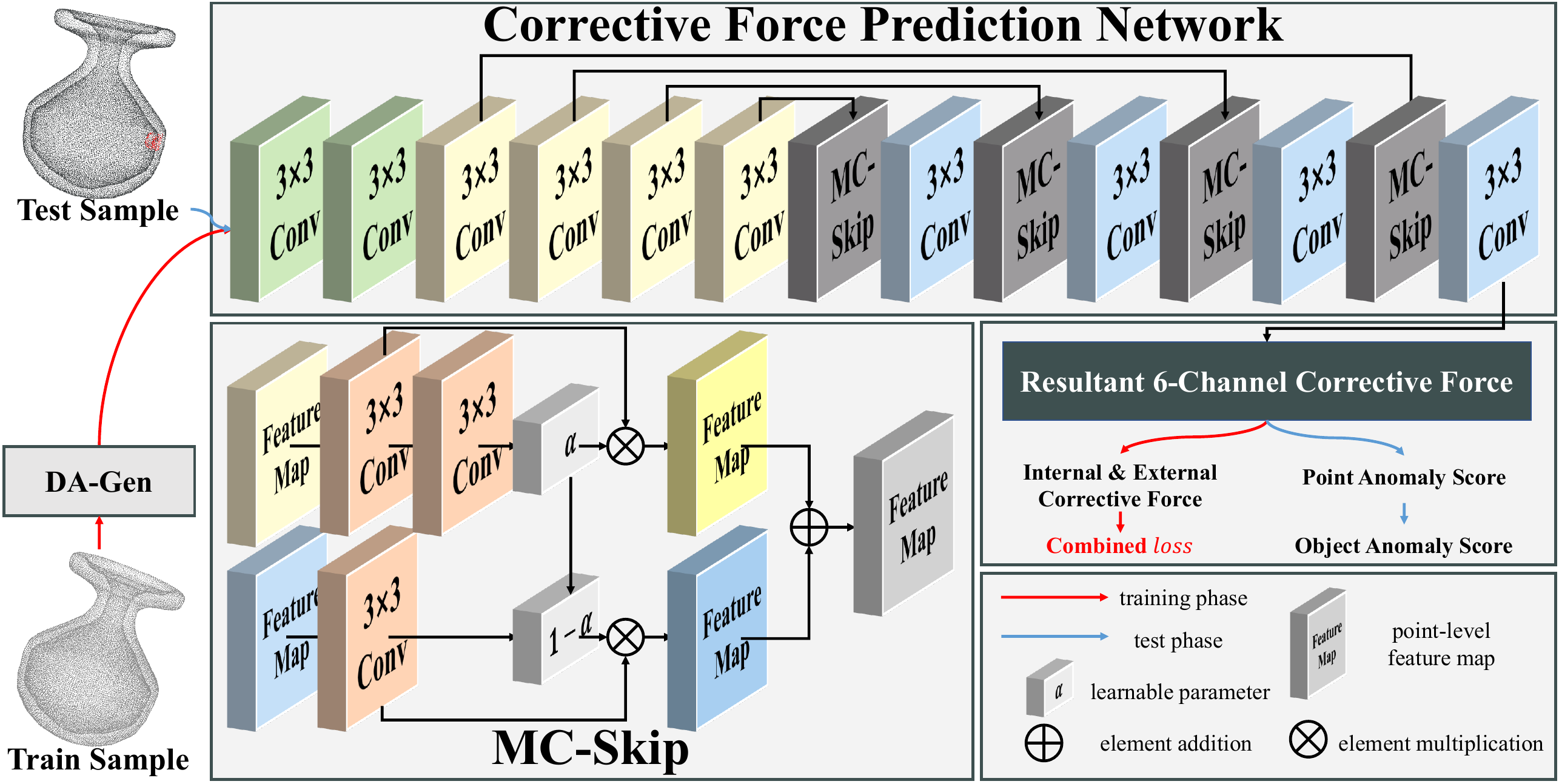}
    \caption{Overview of MC4AD. The proposed MC4AD contains three important parts: 1) \textit{Diverse Anomaly-Generation} (DA-Gen) transfers training normal samples into pseudo-anomaly point clouds. 
        2) \textit{Corrective Force Prediction Network} (CFP-Net) extracts the potential damage forces in the input point cloud and generates the corresponding corrective forces. 
        3) \textit{Combined loss} is used to constrain CFP-Net to generate corrective forces that conform to mechanical constraints correctly.}
    \label{pipeline}
\end{figure}

We propose a novel pipeline for 3D anomaly detection as illustrated in Figure~\ref{pipeline}. 1) Training phrase: The training point cloud is processed by DA-Gen to generate a pseudo-anomaly point cloud and employ CTF-Net to extract the potential damage forces to generate the corresponding corrective forces, a process constrained by the combined loss. 2) Test phrase: CTF-Net processes the test point cloud to generate corrective forces per point, directly converts force magnitudes into point-level anomaly scores, and takes their maximum as the object anomaly score.

\subsection{Diverse Anomaly-Generation}
Effective anomaly elimination in 3D models requires training with realistic pseudo-anomalies that closely resemble real defects in morphology and diversity~\cite{PNAS,PO3AD}. Existing 3D pseudo-anomaly generation methods are limited by their reliance on simplistic geometric patterns and cannot synthesize certain defect types, such as fine scratches. To address this limitation, we introduced Diverse Anomaly-Generation~(DA-Gen) to simulate directional defective forces $F_D$ by perturbing surface normal vectors with controlled randomness. While prior studies indicate that anomalies primarily propagate along normal directions, our method introduces slight directional deviations from these normals to enhance realism. These deviations are constrained to remain proximate to the original normal orientation. Additionally, parametric stretching controls defect elongation, synthesizing narrow non-circular anomalies. Given a point cloud $P$, we partition it into $G$ patches $\hat{P} = \{\hat{P}_i\}_{i=1}^G \in \mathbb{R}^{G \times K \times 3}$, where each patch contains $K$ points sampled via a hybrid strategy combining random selection and $K$-nearest neighbors. The proposed DA-Gen is applied independently to each patch $\hat{P}_i$, formulated as:  
\begin{equation}
    F_{D,i} = \left[ \beta_i \cdot \lambda_i \cdot \nu_i + (1-\lambda_i) \cdot \eta_i \right] \cdot \gamma_i \cdot \frac{\pi_i}{\max(\pi_i)} \cdot \left( 1 - \sigma_i |\pi_i| \right),
\end{equation}
where the subscript $i$ represents the DA-Gen applied to the $i$-th patch. $\beta \in \{-1,1\}$ dictates the displacement direction aligned with or opposing the normal vector $\nu$, $\gamma \in [0.06,0.12]$ specifies the maximum displacement magnitude at the anomaly center, and $\lambda \in [0.95,1]$ governs the dominance of the normal vector $\nu$ over a randomized directional component $\eta$. The normalized projected distance $\pi$ quantifies spatial attenuation of displacements, ensuring nonlinear decay from the center. The stretching parameter $\sigma\in[0,0.08]$ inversely correlates with scratch width. Higher $\sigma$ values yield narrower, more elongated anomalies. Generated samples are reported in Appendix~\ref{Qda}.

\subsection{Corrective Force Prediction Network}
\label{cfp}
Based on the Mechanics Complementary Model, we propose the Corrective Force Prediction Network (CFP-Net), which estimates internal and external corrective forces. Since defective forces constitute continuous vector fields in physical space, our differentiable U-Net architecture explicitly preserves this continuity through its topology. \textbf{By U-Net's universal approximation capability for inverse physical processes, this design directly enables learning implicit defective forces from point clouds and predicting correction forces through rigorous inverse modeling, where its differentiable operators inherently maintain continuous force field embeddings.} 
More analysis of CTP-Net to generate damaging forces and corrective forces is reported in Appendix~\ref{moticfp}.

Our CFP-Net self-adaptively assigns the mechanical contribution of different direction forces in multi-level features by dynamically modulating the complementary feature mechanism. Specifically, our model consists of three modules: an Encoder for extracting potential mechanical features, a Bottleneck for further transforming potential features to target features, and a Decoder for reducing the dimensionality to the effective feature space. Multiple mechanical complementary skip connections~(MC-Skip) are included between the Encoder and Decoder to learn the potential mechanical features. We use $\mathrm{Conv}_i$ to represent $i$-th convolutional layer from MinkUNet~\cite{minkunet}.

Given the $P\text{=}\{\textbf{p}_{i=1}^n\}\in\mathbb{R}^{n\times3}$ mentioned in Section~\ref{motivation}, which employ the map $\varphi$ to voxelize as $V\in \mathbb{R}^{n_V\times3}\ (n_v\leq n)$ and the $n_v$ stands for the voxels number. The Encoder can be formalized as: 
\begin{equation}
\label{encoder}
F_1 = \mathrm{Conv}_1 \big(\varphi(P)\big),\ F_1\in \mathbb{R}^{n_v\times C_1},
\end{equation}
where $\mathrm{Conv}_1$ is two convolutional layers used to map the original point cloud $P$ into $C_1$-dimensional features $F_1$, capturing the differential characterization $F_D(\textbf{p}) = \int_A f_E(\textbf{p}) , dA + \int_{A'} f_I(\textbf{p}) , dA'$ of potential continuous damage source at the point level mentioned in Section~\ref{motivation}.

The features $F_1$ are further processed by a Bottleneck consisting of four convolutional layers to further transform the features to the target feature domain, which can be formalized as: 
\begin{equation}
\label{bottleneck}
F_i = \mathrm{Conv}_i \big( F_{i-1}\big),\ F_i\in\mathbb{R}^{n_v\times C_1},\ \text{for}\ i\in\{2,3,4,5\},
\end{equation}
Three convolutional layers project $C_1$-dimensional features into isomorphic spaces with preserved dimensionality but distinct representations, aligning with the target feature domain.

Establishing dynamic complementary feature learning via Decoder cross-layer integration, where the hierarchical features of the bottleneck are adaptively synthesized with MC-Skip, formulated as:
\begin{equation}
\label{decoder}
\begin{aligned}
F_{11-i} &= \mathrm{Conv}_{10-i}\left(  
    \alpha_{j}*\mathrm{Conv}_{j,1}(F_{10-i})+(1-\alpha_{j})*\mathrm{Conv}_{j,2}(F_{i-1}) 
\right),\ F_{11-i}\in\mathbb{R}^{n_v\times C_{j+1}} ,\ \\
\alpha_{j}&=\mathrm{Conv}_{j,3}(\mathrm{Conv}_{j,2}(F_{i-1})),\ \alpha\in\mathbb{R}^{n_v\times1},\text{for} \ 
  i \in \{5,4,3,2\},\ j \in \{1,2,3,4\},
\end{aligned}
\end{equation}
where $\alpha$ denotes a learnable complementary parameter to balance the different contributions by internal and external features, and the $\mathrm{Conv}_{j,1}$ and $\mathrm{Conv}_{j,2}$ are used to project features to the same feature space. The model generates a feature $F_9\in\mathbb{R}^{n_v\times C_{9}}$ with external and internal coupling characteristics, and the final feature $F_9'$ is assigned to each original point by voxel indexing.

We employ Multilayer Perceptrons~(MLPs) to transform the point-level features $F_9'$, which contain complementary information, into internal and external corrective force vectors that are formalized as:
\begin{equation}
    F_C=MLPs(F_9').\ F_C\in\mathbb{R}^{n\times6},
\end{equation}
where the $F_C$ represents the resultant corrective force mentioned in Section~\ref{motivation}, which is decomposed into $F_E'\in\mathbb{R}^{n\times3}$ and $F_I'\in\mathbb{R}^{n\times3}$, representing the external and internal corrective forces, respectively.

\textbf{Anomaly Score during Inference.} The norm of point-level corrective resultant force $F_C$ is directly considered as an anomaly score during the inference phase. This is attributed to \textbf{Def. A2} in Section~\ref{motivation} that the model gives each point a corrective force that is a zero vector when the fact is normal and a non-zero vector when it is abnormal, as described in Eq.~\ref{eq}. The maximum value of the point-level score is the sample-level score.

\subsection{Combined Loss}
To enhance the reconstruction accuracy of corrective force fields, we propose a combined loss function that integrates three components: a symmetry loss derived from the physical constraint in Eq.~\ref{eq} requiring opposing directions between internal and external correction forces in anomalous regions, a distance loss and a directional consistency loss to optimize global restoration performance jointly. The combinatorial loss function $\mathcal{L}_{comb}$ is defined:
\begin{equation}
\label{loss}
\begin{aligned}
\mathcal{L}_{comb}&=\mathcal{L}_{dist} + \mathcal{L}_{dir}+\mathcal{L}_{sym}= \frac{1}{N}\sum_{i=1}^N\bigg(  \frac{\|F'_{I,i}-F'_{E,i}\|_1}{\|F'_{I,i}\|_2 + \|F'_{E,i}\|_2 + \epsilon} - \frac{F'_{I,i}}{\|F'_{I,i}\|_2+\epsilon}\cdot \frac{F'_{E,i}}{\|F'_{E,i}\|_2+\epsilon} \bigg)\\
 &+\frac{1}{N} \sum_{i=1}^{N}\left\| F_{D,i} - F_{C,i} \right\| - \frac{1}{N} \sum_{i=1}^{N}\frac{F_{D,i}}{\left\| F_{D,i} \right\|_{2} + \epsilon} \cdot \frac{F_{C,i}}{\left\| F_{C,i} \right\|_{2} + \epsilon},\\
\end{aligned}
\end{equation}
where the subscript $i \in \{1,...,N\}$ denotes the $i$-th point in the point cloud and $\epsilon$ is set to 1e-8 to prevent division by zero. This loss enforces anti-symmetric constraints between internal and external forces, motivated by the Eq.~\ref{eq}: For normal points, the resultant corrective force $F_C$ should approximate the zero vector, whereas anomalous points exhibit non-zero resultant forces with complementary force pairs maintaining $\|F'_{E,i}\|_2 = \|F'_{I,i}\|_2$. The distance loss $\mathcal{L}_{dist}$ and the directional consistency loss $\mathcal{L}_{dir}$ to optimize global restoration performance help the model allocate more attention to the anomalous structures.
The combined loss $\mathcal{L}_{comb}$ was used to constrain the overall reconstruction while satisfying our mechanical motivation with the reconstruction effect.

\subsection{3D Anomaly Detection in Industry}

Quality Control (QC) under cost constraints is vital in industrial settings~\cite{qc}, with enterprises facing two challenges: \textit{C1}: multi-level QC using coarse-to-fine detection, and \textit{C2}, adaptive production of varying subcategories. We propose a Hierarchical QC (HQC) framework via three-way decision theory, combining a pruned model with our core method for \textit{C1}, and introduce the Anomaly-IntraVariance dataset to address \textit{C2}’s dynamic conditions. Dataset details are reported in Appendix~\ref{dataintro}.

We implement a two-stage hierarchical detection framework through aggressive model compression: (1) The Bottleneck layers are reduced from three convolutional blocks to two, and the Decoder from four to two blocks, achieving 60\% parameter reduction (2) During inference, the compressed model $V_P$ computes pruning scores $S_P=\{s_i\}_{i=1}^N$ following Section~\ref{cfp}, where samples with $s_i \in \text{Top-}b\%(S_P)$ are classified as normal via thresholding in \textbf{S1}. The remaining $(1-b\%)$ samples undergo full analysis by the original model $V_O$, formalized as:
\begin{equation}
\begin{cases} 
\mathcal{D}_{\text{normal}} = \{P_i \mid \text{rank}(s_i) \leq bN,\ s_i = V_P(P_i)\}, \\
\mathcal{D}_{\text{test}} = V_O(\{P_i\} \setminus \mathcal{D}_{\text{normal}}),
\end{cases}
\end{equation}
where $\text{rank}(s_i)$ denotes the ascending order ranking of pruning scores. Moreover, this HQC can be generalised to the memory bank approach. We describe how to generalise HQC to the memory bank approach and details about the pruned version in Appendix~\ref {gener} and \ref{pruned}.
\section{Experiment}
\subsection{Implementation}
\label{imp}
\textbf{Datasets.}
(1) The \textbf{Anomaly-ShapeNet}~\cite{Li_2024_CVPR} has over 1,600 positive and negative samples from 40 categories, leading to a more challenging setting due to the large number of categories. (2) The \textbf{Anomaly-ShapeNet-New}~\cite{Li_2024_CVPR} provides over 450 positive and negative samples crossing 12 categories, which is not used as an evaluation dataset before. (3) The \textbf{MvTec3D-AD}~\cite{Bergmann_2022} includes 4,147 RGB-D sample pairs from 10 categories, of which 894 are anomalous. We use the depth sample for detection evaluation. (4) The \textbf{Real3D-AD}~\cite{liu2023real3d} dataset consists of 1,254 large-scale, high-resolution point cloud samples from 12 categories, with the training set for each category only containing 4 normal samples. (5) We introduce the \textbf{Anomaly-IntraVariance} dataset, the first 3D dataset for simulating real-world production situations where an assembly line produces multiple products, and contains more than 1,000 samples from 16 categories. We report more details in the Appendix~\ref{dataintro}. 

\textbf{Baselines.} We selected BTF~\cite{BTF}, PatchCore~\cite{9879738}, M3DM~\cite{m3dm}, CPMF~\cite{cao2023CPMF}, Reg3D-AD~\cite{liu2023real3d}, IMRNet~\cite{Li_2024_CVPR}, R3D-AD~\cite{r3dad}, ISMP~\cite{ismp} and PO3AD~\cite{PO3AD} for comparison. 
We report experimental Details, how to choose baselines, and implemented baselines in Appendix~\ref{setting}, \ref{base}, and \ref{impb}, respectively.

\textbf{Evaluation Metrics.} We adopted Point-Level Area Under the Receiver Operator Curve~(P-AUROC, $\uparrow$) and Point-Level Area Under the Per-Region-Overlap~(P-AUPR, $\uparrow$) to evaluate pixel-level anomaly segmentation precision and Object-Level Area Under the Receiver Operator Curve~(O-AUROC, $\uparrow$) and Object-Level Area Under the Per-Region-Overlap~(O-AUPR, $\uparrow$) to evaluate object-level anomaly detection performance. The \textit{Mean Ranking} ($\downarrow$) is also used for evaluation to avoid a small number of categories dominating the average performance. Frames Per Second~(FPS, $\uparrow$) and Parameters~($\downarrow$) are introduced to evaluate the inference speed and size of methods.

\subsection{Main Experiments}
\textbf{Comparison on Anomaly-ShapeNet.} 
Tables~\ref{O-Anomaly} and \ref{P-Anomaly} provide the detection and segmentation results of different methods on Anomaly-ShapeNet across 40 categories, respectively. The proposed approach obtained an outstanding average O-AUROC and P-AUROC of 90.9\% and 91.0\%, outperforming the second-best methods by 7.0\% and 1.2\%, respectively. Moreover, our method reaches 94.0\% of the average results and outperforms the second-best methods by 5.9\%. Our method achieved the best average ranking on O-AUROC, P-AUROC, and O-AURP, demonstrating that our approach outperforms our competitors in most categories.

\begin{table*}[ht]
  \centering
  \caption{The O-AUROC performance ($\uparrow$) of different methods on Anomaly-ShapeNet, where best and second-place results are highlighted in \textcolor[rgb]{1, 0, 0}{\textbf{red}} and \textcolor[rgb]{0, .439, .753}{\textbf{blue}}, respectively. The full version is in Appendix~\ref{moreex}}
  \label{Or-Anomaly2}
  \resizebox{1\textwidth}{!}{
    \begin{tabular}{c|ccccccccccccccccc}
    \toprule
    \textbf{Method} & \textbf{ashtray0} & \textbf{bag0} & \textbf{bottle} & \textbf{bowl} & \textbf{bucket} & \textbf{cap} & \textbf{cup} & \textbf{eraser} & \textbf{headset} & \textbf{helmet} & \textbf{jar} & \textbf{micro.} & \textbf{shelf} & \textbf{tap0} & \textbf{vase0} & \textbf{Average} & \textbf{Mean Rank} \\
    \midrule
    \textbf{BTF(Raw)} & 0.578  & 0.410  & 0.558  & 0.470  & 0.469  & 0.554  & 0.462  & 0.525  & 0.447  & 0.501  & 0.420  & 0.563  & 0.164  & 0.549  & 0.517  & 0.493  & 8.73  \\
    \textbf{BTF(FPFH)} & 0.420  & 0.546  & 0.404  & 0.581  & 0.517  & 0.553  & 0.598  & 0.719  & 0.505  & 0.611  & 0.424  & 0.671  & 0.609  & 0.553  & 0.464  & 0.528  & 8.03  \\
    \textbf{M3DM} & 0.577  & 0.537  & 0.584  & 0.579  & 0.405  & 0.586  & 0.548  & 0.627  & 0.597  & 0.525  & 0.441  & 0.357  & 0.564  & \textcolor[rgb]{ 0,  .439,  .753}{\textbf{0.747 }} & 0.534  & 0.552  & 7.83  \\
    \textbf{PatchCore(FPFH)} & 0.587  & 0.571  & 0.614  & 0.558  & 0.510  & 0.597  & 0.593  & 0.657  & 0.610  & 0.485  & 0.472  & 0.388  & 0.494  & 0.760  & 0.554  & 0.568  & 7.35  \\
    \textbf{PatchCore(PointMAE)} & 0.591  & 0.601  & 0.588  & 0.547  & 0.577  & 0.597  & 0.583  & 0.677  & 0.609  & 0.518  & 0.483  & 0.488  & 0.523  & 0.498  & 0.582  & 0.562  & 7.40  \\
    \textbf{CPMF} & 0.353  & 0.643  & 0.469  & 0.679  & 0.542  & 0.568  & 0.498  & 0.689  & 0.551  & 0.535  & 0.610  & 0.509  & 0.685  & 0.528  & 0.514  & 0.559  & 7.35  \\
    \textbf{Reg3D-AD} & 0.597  & 0.706  & 0.569  & 0.548  & 0.681  & 0.687  & 0.524  & 0.343  & 0.574  & 0.532  & 0.592  & 0.414  & \textcolor[rgb]{ 0,  .439,  .753}{\textbf{0.688 }} & 0.659  & 0.576  & 0.572  & 7.45  \\
    \textbf{IMRNet} & 0.671  & 0.660  & 0.631  & 0.676  & 0.676  & 0.721  & 0.700  & 0.548  & 0.698  & 0.613  & 0.780  & 0.755  & 0.603  & 0.686  & 0.629  & 0.661  & 4.95  \\
    \textbf{R3D-AD} & \textcolor[rgb]{ 0,  .439,  .753}{\textbf{0.833 }} & 0.720  & 0.750  & 0.751  & 0.720  & 0.744  & 0.767  & 0.890  & 0.767  & 0.703  & 0.838  & 0.762  & \textcolor[rgb]{ 1,  0,  0}{\textbf{0.696 }} & 0.818  & 0.734  & 0.749  & 3.23  \\
    \textbf{PO3AD} & \textcolor[rgb]{ 1,  0,  0}{\textbf{1.000 }} & \textcolor[rgb]{ 0,  .439,  .753}{\textbf{0.833 }} & \textcolor[rgb]{ 0,  .439,  .753}{\textbf{0.920 }} & \textcolor[rgb]{ 0,  .439,  .753}{\textbf{0.883 }} & \textcolor[rgb]{ 0,  .439,  .753}{\textbf{0.820 }} & \textcolor[rgb]{ 0,  .439,  .753}{\textbf{0.843 }} & \textcolor[rgb]{ 0,  .439,  .753}{\textbf{0.852 }} & \textcolor[rgb]{ 0,  .439,  .753}{\textbf{0.995 }} & \textcolor[rgb]{ 0,  .439,  .753}{\textbf{0.866 }} & \textcolor[rgb]{ 0,  .439,  .753}{\textbf{0.864 }} & \textcolor[rgb]{ 0,  .439,  .753}{\textbf{0.866 }} & \textcolor[rgb]{ 0,  .439,  .753}{\textbf{0.776 }} & 0.573  & 0.713  & \textcolor[rgb]{ 0,  .439,  .753}{\textbf{0.826 }} & \textcolor[rgb]{ 0,  .439,  .753}{\textbf{0.839 }} & \textcolor[rgb]{ 0,  .439,  .753}{\textbf{2.35 }} \\
    \bottomrule
    \textbf{Ours} & \textcolor[rgb]{ 1,  0,  0}{\textbf{1.000 }} & \textcolor[rgb]{ 1,  0,  0}{\textbf{0.976 }} & \textcolor[rgb]{ 1,  0,  0}{\textbf{0.971 }} & \textcolor[rgb]{ 1,  0,  0}{\textbf{0.943 }} & \textcolor[rgb]{ 1,  0,  0}{\textbf{0.922 }} & \textcolor[rgb]{ 1,  0,  0}{\textbf{0.876 }} & \textcolor[rgb]{ 1,  0,  0}{\textbf{0.964 }} & \textcolor[rgb]{ 1,  0,  0}{\textbf{1.000 }} & \textcolor[rgb]{ 1,  0,  0}{\textbf{0.937 }} & \textcolor[rgb]{ 1,  0,  0}{\textbf{0.930 }} & \textcolor[rgb]{ 1,  0,  0}{\textbf{0.914 }} & \textcolor[rgb]{ 1,  0,  0}{\textbf{0.924 }} & \textcolor[rgb]{ 1,  0,  0}{\textbf{0.696 }} & \textcolor[rgb]{ 1,  0,  0}{\textbf{0.786 }} & \textcolor[rgb]{ 1,  0,  0}{\textbf{0.889 }} & \textcolor[rgb]{ 1,  0,  0}{\textbf{0.909 }} & \textcolor[rgb]{ 1,  0,  0}{\textbf{1.08 }} \\
    \bottomrule
    \end{tabular}%
  }
\end{table*}

\textbf{Comparison on Real3D-AD.}
Tables~\ref{O-Real} and \ref{P-Real} present the detection and segmentation results of different methods on Real3D-AD across 12 categories, respectively. Our model obtained O-AUROC and P-AUROC of 78.6\% and 83.7\%, respectively, maintaining excellent segmentation performance while achieving the SOTA detection performance. This may be because the model predicts better when the training and testing sets vary significantly.

\begin{table*}[ht]
  \centering
  \caption{The O-AUROC performance ($\uparrow$) of different methods on Real3D-AD across 12 categories, where best and second-place results are highlighted in \textcolor[rgb]{1, 0, 0}{\textbf{red}} and \textcolor[rgb]{0, .439, .753}{\textbf{blue}}, respectively.}
  \label{O-Real}
  \resizebox{1\textwidth}{!}{
     \begin{tabular}{c|cccccccccccccc}
    \toprule
    \textbf{Method} & \textbf{Airplane} & \textbf{Car} & \textbf{Candy} & \textbf{Chicken} & \textbf{Diamond} & \textbf{Duck} & \textbf{Fish} & \textbf{Gemstone} & \textbf{Seahorse} & \textbf{Shell} & \textbf{Starfish} & \textbf{Toffees} & \textbf{Average} & \textbf{Mean Rank} \\
    \midrule
    \textbf{BTF(Raw) (CVPR23')} & 0.730  & 0.647  & 0.539  & 0.789  & 0.707  & 0.691  & 0.602  & 0.686  & 0.596  & 0.396  & 0.530  & 0.703  & 0.635  & 7.583  \\
    \textbf{BTF(FPFH) (CVPR23')} & 0.520  & 0.560  & 0.630  & 0.432  & 0.545  & 0.784  & 0.549  & 0.648  & \textcolor[rgb]{ 1,  0,  0}{\textbf{0.779 }} & 0.754  & 0.575  & 0.462  & 0.603  & 7.833  \\
    \textbf{M3DM (CVPR23')} & 0.434  & 0.541  & 0.552  & 0.683  & 0.602  & 0.433  & 0.540  & 0.644  & 0.495  & 0.694  & 0.551  & 0.450  & 0.552  & 9.833  \\
    \textbf{PatchCore(FPFH) (CVPR22')} & \textcolor[rgb]{ 1,  0,  0}{\textbf{0.882 }} & 0.590  & 0.541  & \textcolor[rgb]{ 0,  .439,  .753}{\textbf{0.837 }} & 0.574  & 0.546  & 0.675  & 0.370  & 0.505  & 0.589  & 0.441  & 0.565  & 0.593  & 8.417  \\
    \textbf{PatchCore(PointMAE) (CVPR22')} & 0.726  & 0.498  & 0.663  & 0.827  & 0.783  & 0.489  & 0.630  & 0.374  & 0.539  & 0.501  & 0.519  & 0.585  & 0.594  & 8.750  \\
    \textbf{CPMF (PR24')} & 0.701  & 0.551  & 0.552  & 0.504  & 0.523  & 0.582  & 0.558  & 0.589  & 0.729  & 0.653  & 0.700  & 0.390  & 0.586  & 8.833  \\
    \textbf{Reg3D-AD (NeurIPS23')} & 0.716  & 0.697  & 0.685  & \textcolor[rgb]{ 1,  0,  0}{\textbf{0.852 }} & 0.900  & 0.584  & \textcolor[rgb]{ 0,  .439,  .753}{\textbf{0.915 }} & 0.417  & \textcolor[rgb]{ 0,  .439,  .753}{\textbf{0.762 }} & 0.583  & 0.506  & \textcolor[rgb]{ 0,  .439,  .753}{\textbf{0.827 }} & 0.704  & 5.500  \\
    \textbf{IMRNet (CVPR24')} & 0.762  & \textcolor[rgb]{ 0,  .439,  .753}{\textbf{0.711 }} & 0.755  & 0.780  & \textcolor[rgb]{ 0,  .439,  .753}{\textbf{0.905 }} & 0.517  & 0.880  & 0.674  & 0.604  & 0.665  & 0.674  & 0.774  & 0.725  & 5.000  \\
    \textbf{R3D-AD (ECCV24')} & 0.772  & 0.696  & 0.713  & 0.714  & 0.685  & \textcolor[rgb]{ 1,  0,  0}{\textbf{0.909 }} & 0.692  & 0.665  & 0.720  & \textcolor[rgb]{ 1,  0,  0}{\textbf{0.840 }} & 0.701  & 0.703  & 0.734  & 4.750  \\
    \textbf{ISMP (AAAI25')} & 0.858  & \textcolor[rgb]{ 1,  0,  0}{\textbf{0.731 }} & \textcolor[rgb]{ 1,  0,  0}{\textbf{0.852 }} & 0.714  & \textcolor[rgb]{ 1,  0,  0}{\textbf{0.948 }} & 0.712  & \textcolor[rgb]{ 1,  0,  0}{\textbf{0.945 }} & 0.468  & 0.729  & 0.623  & 0.660  & \textcolor[rgb]{ 1,  0,  0}{\textbf{0.842 }} & 0.757  & \textcolor[rgb]{ 0,  .439,  .753}{\textbf{3.917 }} \\
    \textbf{PO3AD (CVPR25')} & 0.804  & 0.654  & 0.785  & 0.686  & 0.801  & 0.820  & 0.859  & \textcolor[rgb]{ 0,  .439,  .753}{\textbf{0.693 }} & 0.756  & 0.800  & \textcolor[rgb]{ 0,  .439,  .753}{\textbf{0.758 }} & 0.771  & \textcolor[rgb]{ 0,  .439,  .753}{\textbf{0.765 }} & 4.167  \\
    \bottomrule
    \textbf{Ours} & \textcolor[rgb]{ 0,  .439,  .753}{\textbf{0.871 }} & 0.684  & \textcolor[rgb]{ 0,  .439,  .753}{\textbf{0.811 }} & 0.701  & 0.836  & \textcolor[rgb]{ 0,  .439,  .753}{\textbf{0.824 }} & 0.890  & \textcolor[rgb]{ 1,  0,  0}{\textbf{0.701 }} & 0.754  & \textcolor[rgb]{ 0,  .439,  .753}{\textbf{0.803 }} & \textcolor[rgb]{ 1,  0,  0}{\textbf{0.770 }} & 0.785  & \textcolor[rgb]{ 1,  0,  0}{\textbf{0.786 }} & \textcolor[rgb]{ 1,  0,  0}{\textbf{3.083 }} \\
    \bottomrule
    \end{tabular}%
  }
\end{table*}

\textbf{Comparison on Anomaly-ShapeNet-New.} Tables~\ref{O-anomaly-new} and \ref{P-anomaly-new} provide the detection and segmentation results of different methods on Anomaly-ShapeNet-New across 12 categories. The MC4AD obtained an outstanding average O-AUROC and P-AUROC of 88.8\% and 93.7\%, respectively. 

\textbf{Comparison on MvTec3D-AD.}
Table~\ref{O-MVTEC} provides the detection and segmentation results of different only-3D methods on MvTec3D-AD across 12 categories, respectively. The proposed approach obtained an outstanding average O-AUROC and P-AUROC of 95.4\% and 94.6\%, outperforming the previous method, such as M3DM of 94.5\% and 90.6\%. 
This may be attributed to capitalizing on the reason for manufacturing process anomalies to capture structural anomalies. 

\textbf{Comparison on Anomaly-IntraVariance.}
Table~\ref{intra} presents the detection and segmentation results of different methods on Anomaly-IntraVariance across 16 categories, where Group 1 includes two subspecies and Group 2 has four subspecies. Our model achieves 76.1\% and 80.7\% on Group 1 and obtains 62.5\% and 67.8\% on the more challenging Group 2. The average abnormal detection metrics performed 4.40\% and 2.65\% better than the second-best method. 

\subsection{Ablation and Parameter Sensitivity Experiments}
We conducted extensive experiments on Real3D-AD across 12 categories, as shown in Table~\ref{Ablation}.

\begin{table*}[ht]
  \centering
  \caption{Ablation and parameter sensitivity experiments.}
  \label{Ablation}
  \begin{subtable}[t]{0.3\textheight}
    \centering
    \resizebox{\textwidth}{!}{
      \begin{tabular}{c|cccc|>{\columncolor{gray!20}}c>{\columncolor{gray!20}}c}
      \toprule
      Method & $V_1$ & $V_2$ & $V_3$ & $V_4$ & $V_P$ & $V_O$ \\
      \midrule
      $\mathcal{L}_{sym}$ & - & \checkmark & \checkmark & \checkmark & \checkmark & \checkmark \\
      $\mathcal{L}_{dir}$ & \checkmark & - & \checkmark & \checkmark & \checkmark & \checkmark \\
      $\mathcal{L}_{dist}$ & \checkmark & \checkmark & - & \checkmark & \checkmark & \checkmark \\
      $\mathcal{A}_{R}$   & \checkmark & \checkmark & \checkmark & - & \checkmark & \checkmark \\
      \midrule
      O-AUROC($\uparrow$) &0.652&0.678&0.702&0.764&0.754&\textbf{0.786}\\
      P-AUROC($\uparrow$) &0.688&0.705&0.733&0.812&0.792&\textbf{0.837}\\
      \bottomrule
      \end{tabular}%
    }
  \end{subtable}
  \begin{subtable}[t]{0.274\textheight}
    \centering
    \resizebox{\textwidth}{!}{
      \begin{tabular}{l|rrr}
      \toprule
      Method & O-AUROC($\uparrow$)&P-AUROC($\uparrow$)&FPS($\uparrow$) \\
      \midrule
      $w/o_{MC-Skip}$ &\multicolumn{1}{c}{0.733}&\multicolumn{1}{c}{0.740}&\multicolumn{1}{c}{17.1}\\
      $Out=3$&\multicolumn{1}{c}{0.746}&\multicolumn{1}{c}{0.795}&\multicolumn{1}{c}{14.8}\\
      $Layer_2$  &\multicolumn{1}{c}{0.728}&\multicolumn{1}{c}{0.713}&\multicolumn{1}{c}{18.2}\\
      $Layer_4$  &\multicolumn{1}{c}{0.763}&\multicolumn{1}{c}{0.766}&\multicolumn{1}{c}{16.6}\\
      $Layer_6$  &\multicolumn{1}{c}{0.780}&\multicolumn{1}{c}{0.802}&\multicolumn{1}{c}{16.0}\\
      \rowcolor{gray!20}$Layer_8$  &\multicolumn{1}{c}{\textbf{0.786}}&\multicolumn{1}{c}{\textbf{0.837}}&\multicolumn{1}{c}{14.8}\\
      \bottomrule
      \end{tabular}
    }
  \end{subtable}
\end{table*}

\textbf{Analysis of Multiple Loss Functions Effectiveness.} ``$V_1$'' is to predict a point cloud that lacks symmetry constraints, resulting in a predicted imbalance of internal and external Corrective forces. Without $\mathcal{L}_{sym}$, the O-AUROC and P-AUROC decreased by 13.4\% and 14.9\%, respectively, which may be attributed to the force imbalance resulting in an incorrect offset to the normal point. Moreover, we designed ``$V_2$'', where the model is supervised without $\mathcal{L}_{dir}$. 
The performance causes a degradation of the 10.8\% and 13.2\% anomaly detection metrics because the direction of the incorrect force is not constrained, resulting in the direction of the offset not being taken into account in the prediction of the offset. 
Analogous to ``$V_2$'', ``$V_3$'' is proposed for constraining the correct size of the offset. The model performs 8.4\% and 10.4\% decrease in O-AUROC and P-AUROC, respectively, which may be attributed that the model meets challenges to learn the offset distance for both normal and pseudo-abnormal points without $\mathcal{L}_{dist}$. Therefore, all the proposed loss functions are crucial for predicting the defective force and thereby correctly predicting the corrective forces.

\textbf{Analysis of Modules Effectiveness.} In ``$V_4$'', it is observed that O-AUROC and P-AUROC are reduced by 2.2\% and 2.5\%, respectively, due to the replacement of our proposed anomaly generation with Norm-AS in PO3AD, which decreases the learned diversity of anomalies, and thus leads to a decrease in anomaly detection performance. In addition, our pruned version of ``$V_P$'' acquires a more efficient 20 FPS~(>14.8 FPS) while maintaining excellent anomaly detection performance, making our HQC strategy potentially possible, as discussed in Section~\ref{ge}. Moreover, the model without MC-Skip~($w/o\ MC-Skip$) modules failed to capture the complementary features of external and internal corrective forces and was not able to generate the resultant corrective force, resulting in a decrease in the anomaly detection metrics by 5.3\% and 9.7\%, respectively. In contrast, the model directly outputs 3-channel offsets~($Out=3$) equivalent to the direct learning of the corrective force, resulting in a decrease of 4.0\% and 4.2\% in P-AUROC and O-AUROC, respectively. Layer evaluations revealed that reducing layers from 8 to 2 decreased anomaly detection metrics by 5.8\% and 12.4\%, prompting the selection of 8 layers for performance-efficiency balance. More parameter sensitivity experiments are reported in the Appendix~\ref{ab}.

\subsection{Generalizability Experiments and Resource Analysis}
\label{ge}
\textbf{Generalizability Experiments.} We apply MC-Skip to the time-tested classical model, including PointNet~\cite{pointnet}, PointNet++~\cite{pointnet2}, PointTransformer~\cite{pointtransformer}, and DGCNN~\cite{dgcnn}, to validate the generalizability of our model. As shown in Table~\ref{gen}, the accuracies with and without normal participation on the classification task~\cite{wu20153dshapenetsdeeprepresentation} increased on average by 0.65\% and 0.6\%, respectively, and the instance average IoU and class average IoU on the segmentation task~\cite{shapenet} increased on average by 0.775\% and 0.675\%, respectively. Moreover, our HQC strategy performs well on existing 3D anomaly detection models and generalizes to memory bank, as reported in Table~\ref{3w}. Our approach resulted in 10.3\% and 15.3\% FPS average inference rate improvements on Anomaly-ShapeNet and Real3D-AD, respectively, across a representative variety of methods. Implementation details are shown in Appendix~\ref{gend}.

\begin{table*}[ht]
  \centering
  \caption{Comparison between one model and our HQC.}
  \label{3w}
  \resizebox{0.93\textwidth}{!}{
    \begin{tabular}{c|ccc|ccc|ccc|ccc}
    \toprule
    \multirow{3}[6]{*}{\textbf{Method}} & \multicolumn{6}{c|}{\textbf{Anomaly-ShapeNet}} & \multicolumn{6}{c}{\textbf{Real3D-AD}} \\
\cmidrule{2-13}          & \multicolumn{3}{c|}{\textbf{One Model}} & \multicolumn{3}{c|}{\textbf{HQC}} & \multicolumn{3}{c|}{\textbf{One Model}} & \multicolumn{3}{c}{\textbf{HQC}} \\
\cmidrule{2-13}          & \textbf{O-AUROC($\uparrow$)} & \textbf{P-AUROC($\uparrow$)} & \textbf{FPS($\uparrow$)} & \textbf{O-AUROC($\uparrow$)} & \textbf{P-AUROC($\uparrow$)} & \textbf{FPS($\uparrow$)} & \textbf{O-AUROC($\uparrow$)} & \textbf{P-AUROC($\uparrow$)} & \textbf{FPS($\uparrow$)} & \textbf{O-AUROC($\uparrow$)} & \textbf{P-AUROC($\uparrow$)} & \textbf{FPS($\uparrow$)} \\
    \midrule
    \textbf{BTF} &0.528&0.628&4.8&0.543&0.784&\textbf{6.0} \textcolor[rgb]{ 1,  0,  0}{($1.2\uparrow$)}&0.603&0.733&2.5&0.602&0.735&\textbf{3.4} \textcolor[rgb]{ 1,  0,  0}{($0.9\uparrow$)}\\
    \textbf{M3DM} &0.552&0.616&1.9&0.556&0.617&\textbf{2.2} \textcolor[rgb]{ 1,  0,  0}{($0.3\uparrow$)}&0.552&0.631&1.0&0.584&0.632&\textbf{1.5} \textcolor[rgb]{ 1,  0,  0}{($0.5\uparrow$)}\\
    \textbf{Reg3D-AD} &0.572&0.668&1.4&0.588&0.704&\textbf{1.9} \textcolor[rgb]{ 1,  0,  0}{($0.5\uparrow$)}&0.704&0.705&0.8&0.709&0.712&\textbf{1.6} \textcolor[rgb]{ 1,  0,  0}{($0.8\uparrow$)}\\
    \textbf{ISMP} &0.698&0.691&1.3&0.698&0.718&\textbf{2.0} \textcolor[rgb]{ 1,  0,  0}{($0.6\uparrow$)}&0.767&0.836&0.7&0.765&0.845&\textbf{1.4} \textcolor[rgb]{ 1,  0,  0}{($0.5\uparrow$)}\\
    \rowcolor{gray!20}\textbf{Ours} &0.909&0.910&25.0&0.909&0.910&\textbf{26.4} \textcolor[rgb]{ 1,  0,  0}{($1.4\uparrow$)}&0.786&0.837&14.8&0.790&0.837&\textbf{15.7}\textcolor[rgb]{ 1,  0,  0}{($0.9\uparrow$)}\\
    \bottomrule
    \end{tabular}%
  }
\end{table*}

\begin{table*}[ht]
  \centering
  \caption{Results of Generalizability Experiments and Efficiency Analysis.}
  \begin{subtable}[t]{0.442\textwidth}
    \centering
    \resizebox{\textwidth}{!}{
    \begin{tabular}{c|cc|cc}
    \toprule
    \multirow{2}{*}{\textbf{Method}} & \multicolumn{2}{c|}{\textbf{Classification}~\cite{wu20153dshapenetsdeeprepresentation}} & \multicolumn{2}{c}{\textbf{Segmentation}~\cite{shapenet}} \\
    \cmidrule{2-5} & w/o normal & with normal & Instance IoU & Class IoU \\
    \midrule
    PointNet & 89.2 & 90.6 & 83.7 & 80.4 \\
    \rowcolor{gray!20}PointNet(Ours) & \textbf{90.1} \textcolor[rgb]{ 1,  0,  0}{($0.9\uparrow$)} & \textbf{91.4} \textcolor[rgb]{ 1,  0,  0}{($0.8\uparrow$)} & \textbf{84.8} \textcolor[rgb]{ 1,  0,  0}{($1.1\uparrow$)} & \textbf{81.0} \textcolor[rgb]{ 1,  0,  0}{($0.6\uparrow$)} \\
    PointNet++ & 91.9 & 92.2 & 85.1 & 81.9 \\
    \rowcolor{gray!20}PointNet++(Ours) & \textbf{92.4} \textcolor[rgb]{ 1,  0,  0}{($0.5\uparrow$)} & \textbf{92.4} \textcolor[rgb]{ 1,  0,  0}{($0.2\uparrow$)} & \textbf{85.9} \textcolor[rgb]{ 1,  0,  0}{($0.8\uparrow$)} & \textbf{82.6} \textcolor[rgb]{ 1,  0,  0}{($0.7\uparrow$)} \\
    PointTransformer & 93.7 & 93.9 & 86.6 & 83.7 \\
    \rowcolor{gray!20}PointTransformer(Ours) & \textbf{94.1} \textcolor[rgb]{ 1,  0,  0}{($0.4\uparrow$)} & \textbf{94.4} \textcolor[rgb]{ 1,  0,  0}{($0.5\uparrow$)} & \textbf{87.0} \textcolor[rgb]{ 1,  0,  0}{($0.4\uparrow$)} & \textbf{84.4} \textcolor[rgb]{ 1,  0,  0}{($0.7\uparrow$)} \\
    DGCNN & 93.3 & 93.9 & 85.1 & 82.3 \\
    \rowcolor{gray!20}DGCNN(Ours) & \textbf{94.1}\textcolor[rgb]{ 1,  0,  0}{($0.8\uparrow$)} & \textbf{94.8} \textcolor[rgb]{ 1,  0,  0}{($0.9\uparrow$)} & \textbf{85.9} \textcolor[rgb]{ 1,  0,  0}{($0.8\uparrow$)} & \textbf{83.0} \textcolor[rgb]{ 1,  0,  0}{($0.7\uparrow$)} \\
    \bottomrule
    \end{tabular}%
    }
    \caption{Generalization on Classical Models.}
    \label{gen}
  \end{subtable}
  \hfill
  \begin{subtable}[t]{0.482\textwidth}
    \centering
    \resizebox{\textwidth}{!}{
    \begin{tabular}{c|cc|cc}
    \toprule
    \multirow{2}{*}{\textbf{Datasets}} & \multicolumn{2}{c|}{\textbf{Training}} & \multicolumn{2}{c}{\textbf{Inference}} \\
    \cmidrule{2-5} & Frame & Memory & Frame & Memory \\
    \midrule
    
    Anomaly-ShapeNet (PO3AD) & 8.3 & 5.1 & 3.4 & 3.2 \\
    \rowcolor{gray!20}Anomaly-ShapeNet (Ours) & \textbf{30.1} \textcolor[rgb]{ 1,  0,  0}{($21.8\uparrow$)} & \textbf{2.5}  \textcolor[rgb]{ 0,  .439,  .753}{($2.6\downarrow$)} & \textbf{25.0} \textcolor[rgb]{ 1,  0,  0}{($21.6\uparrow$)} & \textbf{1.3} \textcolor[rgb]{ 0,  .439,  .753}{($1.9\downarrow$)} \\

    Anomaly-ShapeNet-New (PO3AD) & 14.2 & 3.1 & 10.8 & 2.8 \\
    \rowcolor{gray!20}Anomaly-ShapeNet-New (Ours) & \textbf{45.5} \textcolor[rgb]{ 1,  0,  0}{($31.3\uparrow$)} & \textbf{2.2} \textcolor[rgb]{ 0,  .439,  .753}{($0.9\downarrow$)} & \textbf{28.1} \textcolor[rgb]{ 1,  0,  0}{($17.3\uparrow$)} & \textbf{2.1} \textcolor[rgb]{ 0,  .439,  .753}{($0.7\downarrow$)} \\

    Mvtec3D-AD (AST) & 49.0 & 11.8 & 50.4 & 5.8 \\
    \rowcolor{gray!20}Mvtec3D-AD (Ours) & \textbf{50.6} \textcolor[rgb]{ 1,  0,  0}{($1.6\uparrow$)} & \textbf{3.9} \textcolor[rgb]{ 0,  .439,  .753}{($7.9\downarrow$)}  & \textbf{50.8} \textcolor[rgb]{ 1,  0,  0}{($0.4\uparrow$)} & \textbf{1.9} \textcolor[rgb]{ 0,  .439,  .753}{($3.9\downarrow$)}  \\

    Real3D-AD (PO3AD) & 7.6 & 6.8 & 3.2 & 5.4 \\
    \rowcolor{gray!20}Real3D-AD (Ours) & \textbf{15.3} \textcolor[rgb]{ 1,  0,  0}{($7.7\uparrow$)} & \textbf{4.4} \textcolor[rgb]{ 0,  .439,  .753}{($2.4\downarrow$)} & \textbf{14.8} \textcolor[rgb]{ 1,  0,  0}{($11.6\uparrow$)} & \textbf{2.9} \textcolor[rgb]{ 0,  .439,  .753}{($2.5\downarrow$)} \\
    \bottomrule
    \end{tabular}%
    }
    \caption{Resource Utilization.}
    \label{resource}
  \end{subtable}
\end{table*}

\textbf{Resource Analysis.}
As shown in Table~\ref{resource}, we report the resource consumption on different datasets and compare it with the model that consumes the minimum resources. In the four datasets available, our rates at training and testing were improved by an average of 15.6 FPS and 12.725 FPS, respectively, compared to the most efficient models available. They only took an average memory of 2.75 GB and 1.075 GB during training and testing, respectively. This may be attributed to the fewer parameters of \textbf{only 14 M}, which results in a significant reduction compared to the PO3AD (34 M). Our model is the preferred choice in industrial anomaly detection, as it is a more efficient method with lower memory.

\begin{figure}[ht]
    \centering
    \includegraphics[width=0.9\linewidth]{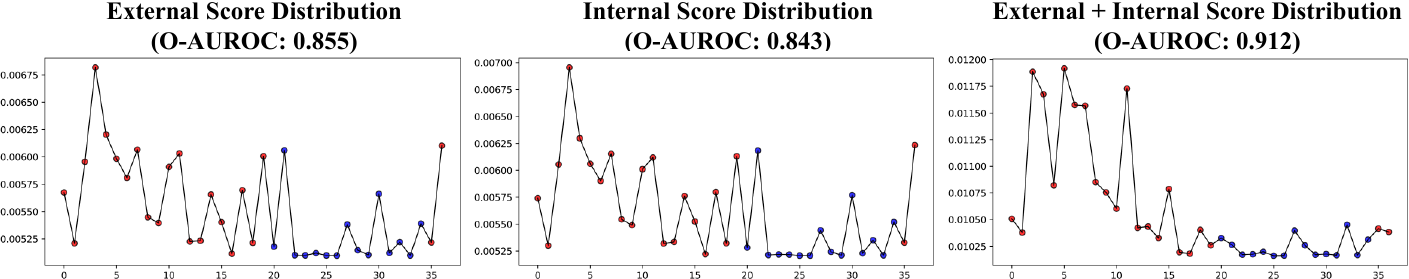}
    \caption{The distribution of anomaly detection scores on class ``vase9'', compared between External, Internal, and Internal+External.  \textcolor[rgb]{1, 0, 0}{\textbf{Red}} and \textcolor[rgb]{0, .439, .753}{\textbf{blue}} represent abnormal and normal points, respectively.}
    \label{s1}
\end{figure}


\begin{wrapfigure}{r}{0.35\textwidth} 
\vspace{-10pt} 
  \centering
  \includegraphics[width=\linewidth]{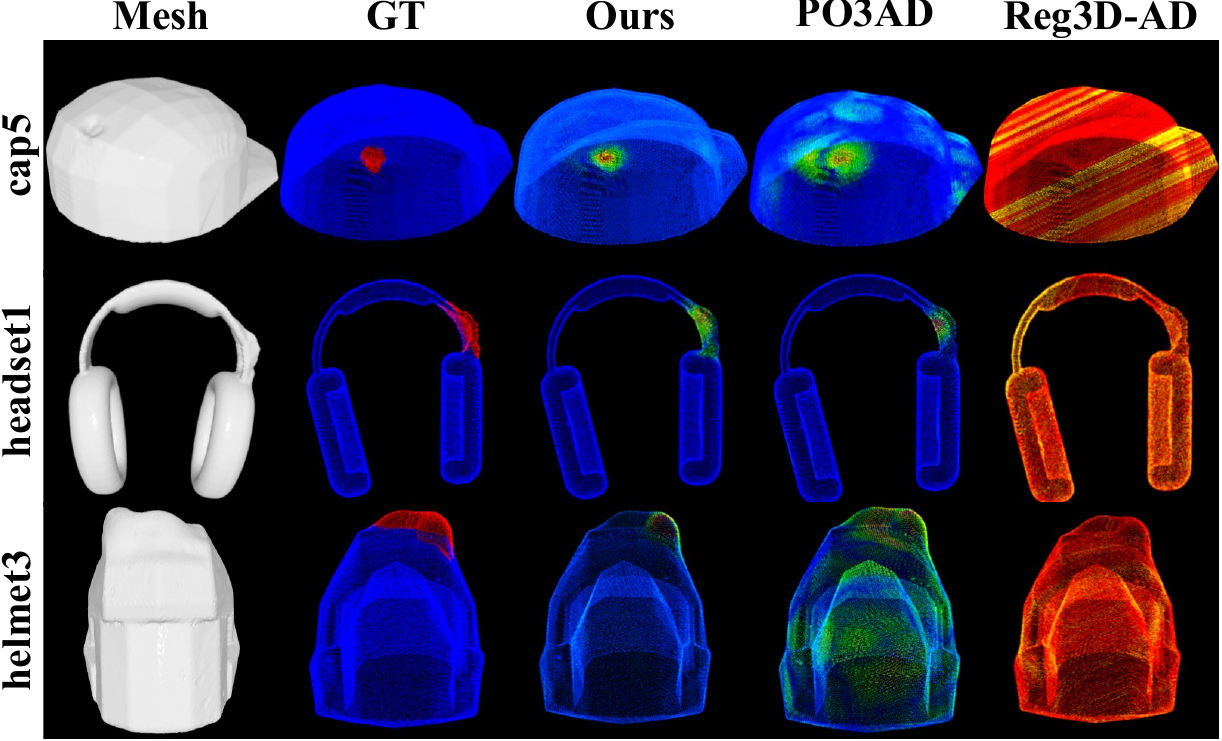}
  \caption{Localization results.}
  \label{vis1}
\vspace{-10pt} 
\end{wrapfigure}

\textbf{Benefit of Combining External \& Internal Corrective Forces.}
MC4AD achieves 90.9\% O-AUROC by integrating bidirectional corrective forces. Figure~\ref{s1} demonstrates that external-internal forces enhance anomaly discrimination by 7.0\% over single-force via complementary feature extraction mechanisms. More comparisons are shown in Appendix~\ref{scoredis}. Our model accurately segments anomalies in complex structures while assigning lower scores to normal parts, as shown in Figure~\ref{vis1}. More results are reported in Appendix~\ref{visdis}.


\section{Conclusion}
This paper presents MC4AD, a novel 3D anomaly detection method that investigates the source of anomalies from a mechanical perspective. 
By considering anomalies as the result of the action of a damaging force, we repaired the anomalies utilizing a corrective force. We generate diverse anomalies by DA-Gen and obtain capture of potentially differentiable damage forces using the proposed CFP-Net to predict the corresponding corrective forces. We also guarantee the correct generation of the corrective force through new combined loss functions. Moreover, a dataset with intraclass variance named Anomaly-IntraVariance and a hierarchical quality control strategy were proposed to simulate a real assembly line environment further. Extensive comparative experiments on the proposed and existing five datasets across 90 categories demonstrate nine advanced anomaly detection and segmentation performance metrics. \textbf{Limitation.} MC4AD lacks explicit physical constraints to derive damage forces for future work; we discuss more in Appendix~\ref{future}.



\clearpage
\appendix

\appendixtitle{Supplementary Material of ``Examining Defects from a Mechanics Perspective for 3D Anomaly Detection''}

\section{Experimental Details}
\subsection{Experimental Settings}
\label{setting}
Our experiments were trained on four NVIDIA A100-PCIE-40GBs and an Xeon Gold 6248R machine. The batch size was set to 32, and the initial learning rate was 0.001 and declined with a cosine annealing scheme~\cite{loshchilov2017sgdrstochasticgradientdescent}. The optimiser chosen was Adam. Model test results were implemented on a machine fitted with an NVIDIA RTX3090 and a 12 vCPU Intel(R) Xeon(R) Platinum 8255C CPU. the number of anomalous blocks created by DA-Gen was 64 and $\lambda$ was set to a random number within [0.95,1], $\sigma$ was set to a random number within [0,0.08], $\gamma$ was set to a random number within [0.06,0.12]. $b$ in HQC was set to 0.25. All samples were taken without using downsampling. All code is implemented using PyTorch.

\subsection{How We Select Baseline}
\label{base}
We select models according to their performance, and some lower-performing models are excluded from the selection.  Moreover,  some excellent work based on large language models has excelled in 3D anomaly detection. Considering the large amount of data they use that is difficult to access in the real industry, we compare models trained with limited data, such as migration training using PointMAE, or local datasets, such as those taught on Real3D-AD. For this reason, we state some models that were not selected for comparison. 

\textbf{GLFM}~\cite{GLFM} uses pre-trained Vision-Language Models (VLMs) as feature extractors.

\textbf{PointAD}~\cite{pointad} employs multi-view projection is used to form 2D images, and 2D-LLM is used to acquire features.

\textbf{ENEL}~\cite{ENEL} uses an end-to-end encoder-free large language model for 3D anomaly detection.

The above work is excellent, and we collected it in Arxiv or the official conference website. We discuss them in related work, and we do not impose comparison experiments due to the task setup.

\subsection{Implementation of Baselines}
\label{impb}
\textbf{BTF} extracts point cloud features using FPFH and creates a memory bank to determine anomalies by comparing the features to be tested with the memory bank features. We conducted comparative experiments using the results from their paper, and we measured the unreported results via the official implementation at the following link: \url{https://github.com/eliahuhorwitz/3D-ADS}. 

\textbf{PatchCore} is a classical approach in 2D anomaly detection, which Real3D-AD extends to 3D anomaly detection. The paper uses different feature extractors~\cite{pointbert,resnet} to obtain features from the point cloud registered by RANSAC~\cite{ransac}. The code implementation is derived from the official Real3D-AD implementation, which can be acquired at the following link: \url{https://github.com/m-3lab/real3d-ad}.

\textbf{M3DM} uses PointMAE to obtain group-level features, which Real3D-AD extends to 3D-only anomaly detection. For the MvTec3D-AD detection results, we used the official implementation of M3DM, which is available at the following link: \url{https://github.com/nomewang/M3DM/blob}. For a 3D-only implementation of the paper, we used the Real3D-AD implementation, which can be accessed at the following link: \url{https://github.com/m-3lab/real3d-ad}.

\textbf{CPMF} uses multi-view projections to render pseudo-2D modalities and ResNet for representations as a complement to 3D features. The code is available at the link: \url{https://github.com/caoyunkang/CPMF}.

\textbf{Reg3D-AD} extracts features from the RANSAC-registered point cloud by using PointMAE~\cite{pointmae}. We get the code from the following link: \url{https://github.com/m-3lab/real3d-ad}.

\textbf{IMRNet} eliminates potential anomalies in the point cloud by expanding PointMAE to mask the reconstruction circularly. Since its code is not open-sourced, we obtained its official paper performance through the following official repository: \url{https://github.com/Chopper-233/Anomaly-ShapeNet}.

\textbf{R3D-AD} eliminates potential anomalies from a point cloud by a diffusion model, and we obtained its official paper performance through the following official repository: 
\url{https://github.com/zhouzheyuan/r3d-ad}.

\textbf{ISMP} obtains information about the interior of the point cloud through pseudo-modal projections as complementary information to get better anomaly detection, and we access the official implementation through the following link: \url{https://github.com/M-3LAB/Look-Inside-for-More}.

\textbf{PO3AD} eliminates potential anomalies from a point cloud by predicting point offsets. Its official implementation can be accessed at the following link: \url{https://github.com/yjnanan/PO3AD}.

\subsection{Details of the HQC Generalisation to Memory Bank Methods}
\label{gener}
The core of HQC is an initial screening with a simplified version that uses its faster but slightly reduced detection performance to detect a fraction of significantly normal samples, leaving the other fraction of potentially abnormal samples for further detection using the original model. When properly initialized, the model speeds up detection without losing accuracy.

Following the settings of Reg3D-AD, ISMP, and PatchCore, whose standard memory bank size is set to 10000. We simplify this by setting it to 1000 to detect anomalies according to the hierarchical detection of HQC.

\subsection{Details of the Pruned Version}
\label{pruned}
We reduce the convolutional layers in Encoder, Bottleneck, and Decoder represented by Eq.~\ref{encoder}, Eq.~\ref{bottleneck}, and Eq.~\ref{decoder} to one, two, and two, respectively. Since the number of convolutions in the bottleneck layer and the Decoder remains the same, the structure of the MC-Skip is retained as in the original version to look for potential corrupting forces in the internal and external structures.

Moreover, MinkUNet is a voxel-based sparse convolutional network. All convolutions used in our model were replaced with the convolutional layers used in MinkUnet, and we achieved standard performance by using the standard configuration from their paper, with the $voxel\ size$ set to 0.03.

The pruned version follows the same training setup as described in Section~\ref{setting}, and since the model structure is simpler compared to the original version, we adjust the initial learning rate upwards to 0.0015 to ensure faster convergence.

\subsection{Details of Generalizability Experiments}
\label{gend}
According to the paper setup, PointNet++, DGCNN, and PointTransformer are based on PointNet. We first conduct generalisation experiments on PointNet, and then plug-and-play into the corresponding classical models according to their thesis settings. The representation of a point cloud $P$ as a feature $F_{pointnet}$ by PointNet can be formalised as:
\begin{equation}
F_{pointnet}=Conv_3\bigg(Conv_2\big(Conv_1\big(T\text{-}Net(P)\big)\big)\bigg),
\end{equation}
Reformalised by inserting MC-Skip as
\begin{equation}
\label{modi}
\begin{aligned}
F_1&=T\text{-}Net(P),\\
F_2&=\mathrm{Conv}_1(F_1),\\
F_3&=\mathrm{Conv}_2(F_2),\\
F_{4} &= \mathrm{Conv}_{3}\left(  
    \alpha_{1}*\mathrm{Conv}_{1,1}(F_{2})+(1-\alpha_{1})*\mathrm{Conv}_{1,2}(F_{3}) 
\right),\\
F_{5} &= \mathrm{Conv}_{4}\left(  
    \alpha_{2}*\mathrm{Conv}_{2,1}(F_{1})+(1-\alpha_{2})*\mathrm{Conv}_{2,2}(F_{4}) 
\right), \\
\alpha_{1}&=\mathrm{Conv}_{1,3}(\mathrm{Conv}_{1,2}(F_{3})),\ \alpha_1\in\mathbb{R}^{n_v\times1},\\
\alpha_{2}&=\mathrm{Conv}_{2,3}(\mathrm{Conv}_{2,2}(F_{4})),\ \alpha_2\in\mathbb{R}^{n_v\times1},
\end{aligned} 
\end{equation}
Based on the above formulation, the process encoded by PointTransformer using PointNet can be reformulated:
\begin{equation}
\begin{aligned}
    \hat{P}_i,C_i&=KNN(FPS(P)),\\
    \hat{F}_i&=PointNet(\hat{P}_i),\\
    F_i&=Attention(\hat{F}_i,C_i).
\end{aligned}
\end{equation}
where $KNN$ and $FPS$ stand for K Nearest Neighbours and Farthest Point Sampling, respectively, and are commonly used to divide a point cloud into patches. $C_i$ and $\hat{P}_i$ represent the center point and group points of the $i$-th patch divided, respectively. Each Patch is encoded using modified PointNet to get the encoded feature $\hat{F}_i$, and $Attention$ is used for final features $F_i$ representing the features of the $i$-th patch.

Following the modified PointNet of Eq.~\ref{modi}, the DGCNN inserts dynamic graph aggregation after each convolution. Moreover, the PointNet++ feature extractor for multi-scale point clouds using PointNet.

\section{More Analysis about Motivation}
\label{moremoti}
In this paper, we view most anomalies in industrial manufacturing as the combined result of internal and external damage forces.
Moreover, we elaborate further on the individual formulas' theoretical implications and relate them to 3D anomaly detection.
\subsection{Analysis of Potential Continuous Damage Force and Correction Force}
Continuous forces are usually expressed as integrals of force densities over a spatial distribution. Mass $M$ can be written as a product of density $\rho$ and volume $V$~\cite{phy}. In 3D-AD, the basic unit for detecting samples is the point cloud. Therefore, we ignore the object's mass density, and each point can be ideally regarded as a unit of density. Furthermore, as the point cloud is composed of discrete points, the volume $V$ can be approximated as a continuous surface $S$. The force representation in the 3D point cloud can be written:
\begin{equation}\label{eq2}
\begin{aligned}
    F &= \int_M f \, dM, \\
      &= \int_V f \, d(\rho V),\\
                   &=\int_V \rho f \, dV,\\
                   &= \int_V f \, dV,\\
                   &= \int_S f \, dS,
\end{aligned}
\end{equation}
where $S$ represents the force acting uniformly on the internal and external surfaces. 

Eq.~\ref{eq2} represents the combined action of complex resultant forces on the surface $S$. However, making the model learn the complex potential resultant forces is straightforward. Therefore, we further decompose the resultant forces to obtain the external force $F_E$ and internal force $F_I$, which can be represented as follows:
\begin{equation}
\begin{aligned}
    F_E &= \int_{S^{+}} f_E \, dS^{+},\\
    F_I &= \int_{S^{-}} f_I \, dS^{-},\\
\end{aligned}
\end{equation}
where $S^+$ and $S^-$ represent the external and the internal part of the surface $S$, respectively.

Considering the above theoretical basis, we formalise the damage force $F_D$ and corrective forces $F_C$ in the paper according to the internal and external decompositions in \textbf{Definition A2} and \textbf{Definition A3}, respectively.

\subsection{Analysis of Different Forces to Normal Points and Anomaly Points}
According to Eq.~\ref{eq}, the damage and corrective forces for a particular point $\textbf{p}$ to be judged as normal can be described as
\begin{equation}\label{eq3}
\begin{aligned}
if\ \textbf{p}\ is\ normal\ F_D(\textbf{p})&=\textbf{0}\ else \neq \textbf{0},\ F_C(\textbf{p})=-F_D(\textbf{p}),
\end{aligned}
\end{equation}
The above equation can be fully expressed as:
\begin{equation}\label{eq4}
\begin{aligned}
if\ \textbf{p}\ is\ normal,\ F_D(\textbf{p})&=F_C(\textbf{p})=\textbf{0},\\
if\ \textbf{p}\ is\ anomaly,\ F_D(\textbf{p})&=-F_C(\textbf{p})\neq \textbf{0},
\end{aligned}
\end{equation}
Eq.~\ref{eq4} can be derived from Newton's First Law of Motion. if the normal structure is undamaged, the point is subjected to zero damaging force, and we do not need to apply a corrective force to rectify it.
On the contrary, if a non-zero damage force is suffered, the point is anomalous and, according to the definition in \textbf{Definition A4}, we need to apply an opposite corrective force to repair it.

\subsection{Theoretical Analysis of the CFP-Net.}
\label{moticfp}
Since our model learns potential damage forces from the point cloud and generates corrective forces to counteract the effects of the damage forces through an inverse process, this process behaves as an asymmetric inverse process. The model is required to solve the following challenges:

 \textcolor[rgb]{0, 0, 1}{\textbf{Challenge A1.}} Inverse Process Capability: Given that the model aims to perform an inverse process, it should be adept at reconstructing corrective forces through inverse operations.

 \textcolor[rgb]{0, 0, 1}{\textbf{Challenge A2.}} Complementary Force Learning: To directly learn internal and external forces at each point and synthesize the resultant force, the model must capture the complementary nature of latent internal and external forces as mentioned in \textbf{Definition A2} and \textbf{Definition A3}, thereby generating accurate resultant corrective forces.

 \textcolor[rgb]{0, 0, 1}{\textbf{Challenge A3.}} Differentiability: Considering the continuity and differentiability of force distributions as mentioned in Appendix~\ref{moremoti}, the proposed model must be differentiable to enable precise mathematical operations such as gradient-based optimization. 

\underline{\textbf{To address the challenges above, our model incorporates the following distinctive features:}}

\textcolor[rgb]{1, 0, 0}{\textbf{For Challenge A1.}} As our model adopts a U-Net framework, it inherently satisfies two fundamental theorems:  

\framebox{\begin{minipage}{\dimexpr\textwidth-2\fboxsep-2\fboxrule}
\textit{Lemma A1~\cite{le1}.} A feedforward neural network with a linear output layer and at least one hidden layer employing any "squashing" activation function can approximate any Borel measurable function from one finite-dimensional space to another with arbitrary precision, provided sufficient hidden units are available.

\quad

\textit{Lemma A2~\cite{le2}.} The U-Net architecture is universally applicable for arbitrary inverse generative tasks in computer vision.  
\end{minipage}}
Given \textit{Lemma A1} and \textit{Lemma A2}, our proposed model theoretically guarantees superior performance in fitting the required inverse process.

\textcolor[rgb]{1, 0, 0}{\textbf{For Challenge A2.}} As explicitly formulated in Eq.~\ref{decoder} through complementary constraints, our model employs the MC-Skip mechanism to dynamically adjust feature weights across different hierarchies. This design guides the model to learn complementary feature representations, which are further regularized by a symmetric loss function as mentioned in Eq.~\ref{loss}.

 \textcolor[rgb]{1, 0, 0}{\textbf{For Challenge A3.}} Our model builds upon the differentiable Minkowski-UNet architecture with further modifications to ensure enhanced differentiability. We prove it simply by the following process.
\framebox{\begin{minipage}{\dimexpr\textwidth-2\fboxsep-2\fboxrule}

\subsubsection*{Proposition A1}
The mapping $\mathcal{F}: \mathbf{P} \mapsto \mathbf{F}_9$ is differentiable with respect to $\mathbf{P}$ and all parameters $\{\mathbf{W}_1, \mathbf{W}_i, \mathbf{W}_{j,k}\}$.

\begin{proof}
\quad\\
\textbf{\emph{Convolutional Differentiability:}}
\begin{equation*}
\begin{aligned}
    \because \quad \frac{\partial \text{Conv}(\mathbf{F}; \mathbf{W})}{\partial \mathbf{F}}& = \mathbf{W}, \\
    \quad \frac{\partial \text{Conv}(\mathbf{F}; \mathbf{W})}{\partial \mathbf{W}} &= \mathbf{F},
\end{aligned}
\end{equation*}
where holds for all encoder/bottleneck layers $\text{Conv}_k$ ($k=1,...,5$).

\textbf{\emph{Dynamic Weight Differentiability:}}
\begin{equation*}
\begin{aligned}
     &\because \quad \alpha_j= \sigma\left( \mathbf{W}_{j,3} \ast (\mathbf{W}_{j,2} \ast \mathbf{F}_{j+1}) \right),\\
    &\therefore \quad \frac{\partial \alpha_j}{\partial \mathbf{F}_{j+1}} = \sigma'\left( \mathbf{W}_{j,3} \ast (\mathbf{W}_{j,2} \ast \mathbf{F}_{j+1}) \right) \cdot (\mathbf{W}_{j,3} \ast \mathbf{W}_{j,2}),
\end{aligned}
\end{equation*}

where $\sigma'(z) = \sigma(z)(1-\sigma(z))$ is continuous.

\textbf{\emph{Decoder Gradient Accumulation:}}

For decoder layer $\mathbf{F}_{10-j}$:
\begin{equation*}
\begin{aligned}
    &\because \quad \mathbf{F}_{10-j} = \mathbf{W}_{10-j} \ast \left( \alpha_j (\mathbf{W}_{j,1} \ast \mathbf{F}_{6-j}) + (1-\alpha_j)(\mathbf{W}_{j,2} \ast \mathbf{F}_{j+1}) \right),\\
    &\therefore \quad \frac{\partial \mathbf{F}_{10-j}}{\partial \mathbf{P}} = \alpha_j \mathbf{W}_{j,1} \frac{\partial \mathbf{F}_{6-j}}{\partial \mathbf{P}} + \left[ (1-\alpha_j)\mathbf{W}_{j,2} + \frac{\partial \alpha_j}{\partial \mathbf{F}_{j+1}}(\mathbf{W}_{j,1}\mathbf{F}_{6-j} - \mathbf{W}_{j,2}\mathbf{F}_{j+1}) \right] \frac{\partial \mathbf{F}_{j+1}}{\partial \mathbf{P}},
\end{aligned}
\end{equation*}

\textbf{\emph{Global Differentiability:}}
\begin{equation*}
\begin{aligned}
    &\because \quad \frac{\partial \Phi}{\partial \mathbf{P}} = \sum_{\text{all paths } p} \prod_{(i \to j) \in p} \frac{\partial \mathbf{F}_j}{\partial \mathbf{F}_i},\\
    &\therefore \quad \text{All terms } \frac{\partial \mathbf{F}_j}{\partial \mathbf{F}_i} = \mathbf{W}_k \text{ or fusion terms are continuous}.
\end{aligned}
\end{equation*}
Thus $\mathcal{F}$ is differentiable by composition of differentiable operations.
\end{proof}
\end{minipage}}
Since the whole process is microscopic, our model is more suitable for learning potentially microscopic variables.

\textbf{In conclusion, we theoretically ensure that the CFP-Net method is designed specifically to extract potential damage forces for 3D anomaly detection.}

\section{Main Symbols}
\label{ms}
We present the main symbols used in the theoretical analysis and corresponding explanations, as shown in Table~\ref{symbol}.

\begin{table}[t!]
\centering
\caption{Main symbols and corresponding explanations}
\label{symbol}
\resizebox{0.9\columnwidth}{!}{%
\begin{tabular}{@{}ll@{}}
\toprule
\textbf{Symbol} & \textbf{Explanation} \\
\midrule
$\textbf{M}$ & A normal sample as 3-manifold \\
$\textbf{M}'$ & A anomaly sample as 3-manifold \\
$P\{\textbf{p}_{i=1}^n\}\in\mathbb{R}^{n\times3}$ & A point cloud with each point $\textbf{p}_i=(x_i,y_i,z_i)^\top$ \\
$F_D(\textbf{p})\in \mathbb{R}^3$ & A resultant defective force \\
$F_E(\textbf{p})\in \mathbb{R}^3$ & A resultant external defective force \\
$F_I\in \mathbb{R}^3$ & A resultant internal defective force \\
$f_D(\textbf{p})$, $f_E(\textbf{p})$, $f_I(\textbf{p})$ & Resultant defective force, component external defective force, and component internal defective force \\
$S^+$ \& $S^{-}$ & External and internal planes \\
$F_C\in \mathbb{R}^3$ & A resultant corrective force \\
$F_E'\in \mathbb{R}^3$ & A resultant external corrective force \\
$F_I'\in \mathbb{R}^3$ & A resultant internal corrective force \\
$f_D'(\textbf{p})$, $f_E'(\textbf{p})$, $f_I'(\textbf{p})$ & Component force, component external corrective force, and component internal corrective force \\
$\phi:(F,G)\rightarrow\nabla G$ & A non-linear additive map representing the affect of the focre $F$ to object $G$ \\
$M$ & Mass \\
$\rho$ & A product of density \\
$V$ & Volume \\
$S$ & A continuous surface \\
\bottomrule
\end{tabular}%
}
\end{table}

\section{Datasets: Anomaly-IntraVariance}
\label{dataintro}
\subsection{Data Overview}
\begin{figure}[!ht]
    \centering
    \includegraphics[width=0.8\linewidth]{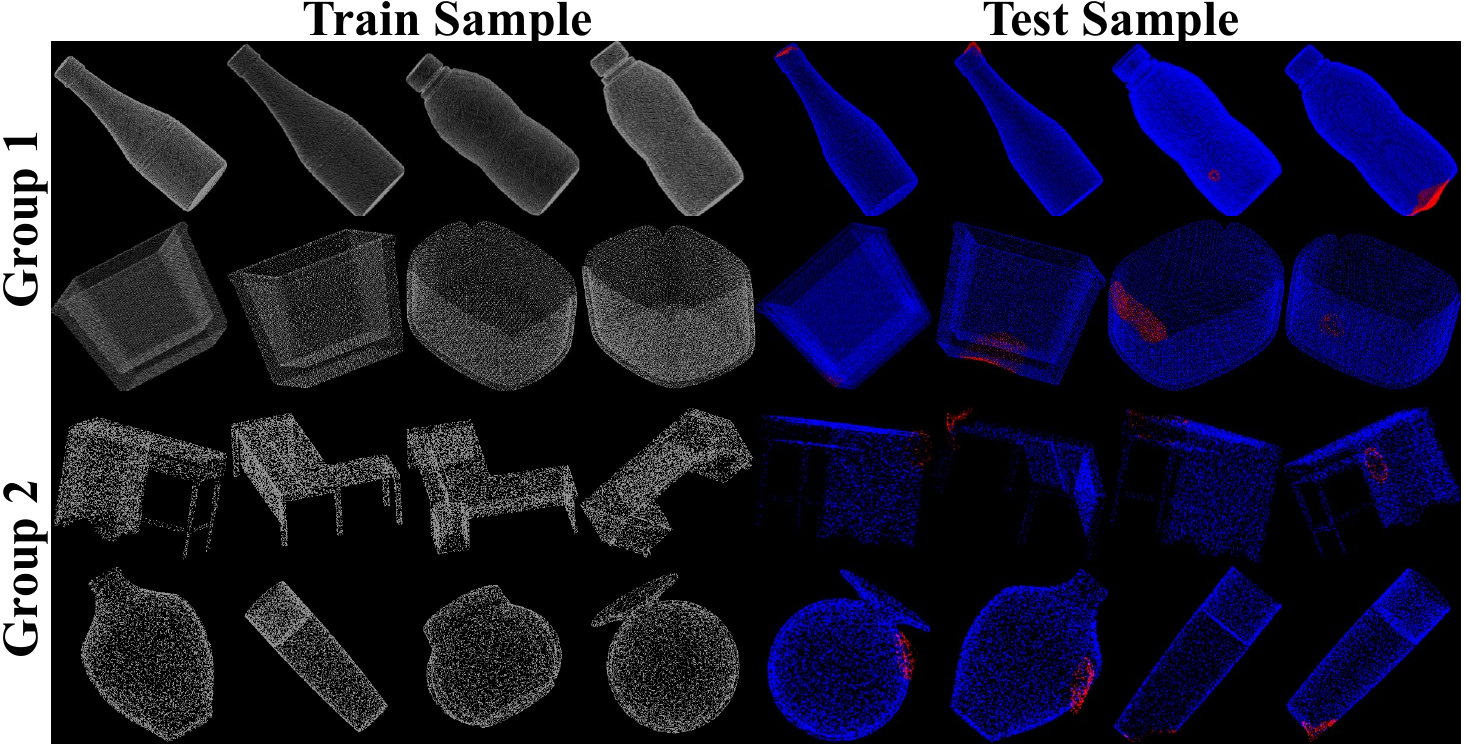}
    \caption{Visualization of the training and test sets on Group 1 and Group 2.}
    \label{vis2}
\end{figure}

We constructed an anomaly detection dataset with intra-class variance as shown in Figure~\ref{vis2}. Group 1 contains eight classes, each class having two subclasses. Group 2 also includes eight classes, where each class has four subclasses. Moreover, to more closely mimic 3D anomaly detection in a real industrial context, Group 2 has fewer points and has been subjected to light noise. This is attributed to the tens of minutes required to scan a high-resolution sample such as Real3D-AD, which may pose challenges in the real industry. Thus, we provide a more comprehensive test benchmark for 3D anomaly detection in the real sector.
\begin{table*}[t!]
  \centering
  \caption{Statistical Data of Group 1.}
  \resizebox{0.9\textwidth}{!}{
    \begin{tabular}{cc|cccccccccc}
    \toprule
    \multicolumn{12}{c}{\textbf{Statistical Data}} \\
    \midrule
    \textbf{class} & \textbf{subcategories} & \textbf{template} & \textbf{positive} & \textbf{hole} & \textbf{bulge} & \textbf{concavity} & \textbf{crak} & \textbf{broken} & \textbf{scratch} & \textbf{bending} & \textbf{total} \\
    \midrule
    \textbf{bottle} & \textbf{2} & 4     & 30    & 2     & 14    & 14    & 1     & 2     & \textbackslash{} & \textbackslash{} & \textbf{67} \\
    \textbf{bowl} & \textbf{2} & 4     & 30    & \textbackslash{} & 14    & 14    & \textbackslash{} & \textbackslash{} & 8     & \textbackslash{} & \textbf{70} \\
    \textbf{bucket} & \textbf{2} & 4     & 30    & 4     & 14    & 14    & 4     & 4     & 2     & \textbackslash{} & \textbf{76} \\
    \textbf{cap} & \textbf{2} & 4     & 30    & 4     & 14    & 14    & \textbackslash{} & 4     & \textbackslash{} & 1     & \textbf{71} \\
    \textbf{cup} & \textbf{2} & 4     & 30    & \textbackslash{} & 14    & 14    & \textbackslash{} & \textbackslash{} & \textbackslash{} & \textbackslash{} & \textbf{62} \\
    \textbf{helmet} & \textbf{2} & 4     & 30    & 2     & 14    & 14    & 2     & 2     & 1     & 2     & \textbf{71} \\
    \textbf{microphone} & \textbf{2} & 4     & 30    & \textbackslash{} & 14    & 14    & \textbackslash{} & \textbackslash{} & \textbackslash{} & \textbackslash{} & \textbf{62} \\
    \textbf{tap} & \textbf{2} & 4     & 30    & 4     & 14    & 14    & 2     & 4     & 2     & \textbackslash{} & \textbf{74} \\
    \bottomrule
    \end{tabular}%
  \label{data1}
    }
\end{table*}
\begin{table*}[t!]
  \centering
  \caption{Statistical Data of Group 2.}
  \resizebox{0.9\textwidth}{!}{
    \begin{tabular}{cc|ccccccc}
    \toprule
    \multicolumn{9}{c}{\textbf{Statistical Data}} \\
    \midrule
    \textbf{class} & \textbf{subcategories} & \textbf{template} & \textbf{positive} & \textbf{missing} & \textbf{bulge} & \textbf{crack} & \textbf{twist} & \textbf{total} \\
    \midrule
    \textbf{cone} & \textbf{4} & 4     & 25    & 11    & 7     & 4     & 8     & \textbf{59} \\
    \textbf{bowl} & \textbf{4} & 4     & 25    & 10    & 7     & 3     & 7     & \textbf{56} \\
    \textbf{door} & \textbf{4} & 4     & 21    & 2     & 4     & 3     & 7     & \textbf{41} \\
    \textbf{keyboard} & \textbf{4} & 4     & 24    & 11    & 8     & 4     & 6     & \textbf{57} \\
    \textbf{night\_stand} & \textbf{4} & 4     & 31    & 5     & 6     & 3     & 6     & \textbf{55} \\
    \textbf{raido} & \textbf{4} & 4     & 25    & 8     & 7     & 4     & 7     & \textbf{55} \\
    \textbf{vase} & \textbf{4} & 4     & 32    & 12    & 8     & 4     & 8     & \textbf{68} \\
    \textbf{xbox} & \textbf{4} & 4     & 29    & 11    & 8     & 3     & 8     & \textbf{63} \\
    \bottomrule
    \end{tabular}%
      \label{data2}
    }
\end{table*}

\subsection{Anomaly Data Synthesis}
We collated samples of Anomaly-ShapeNet and used this to generate Group 1 to produce simpler anomaly detection data.
Following Anomaly-ShapeNet, we selected some of the categories on ShapeNet~\cite{wu20153dshapenetsdeeprepresentation} as base samples to generate Group 2 and performed manual anomaly generation by using Blender software, commonly used in industry. In a further effort to simulate the noise problem present in industry, we imported Gaussian noise with a variance of 0.002 to simulate environmental and equipment effects. Finally, we labelled the ground truth through CloudCompare.

\subsection{Dataset Statistics.}
The data contains more than 1,000 samples in 16 categories, four samples in the training set, and dozens of samples in the test set, and the statistics for Group 1 and Group 2 are shown in Tables~\ref{data1} and \ref{data2}, respectively. Sample point clouds range from 7,000 to 25,000 points, meeting the scanning accuracy of the actual industry. 

\section{More Ablation Results}
\label{ab}

\begin{table*}[!ht]
  \centering
  \caption{More Parameter Sensitivity Analyses.}
  \label{tab:combined}
  \begin{subtable}[t]{0.43\textwidth}
    \centering
    \caption{Parameter Sensitivity Analysis to $G$.}
    \label{a1}
    \resizebox{\textwidth}{!}{
      \begin{tabular}{c|cccc}
      \toprule
      Number & 16 & 32 & \textbf{64} & 128 \\
      \midrule
      O-AUROC & 0.769 & 0.780 & \textbf{0.786} & 0.766 \\
      P-AUROC & 0.810 & 0.812 & \textbf{0.837} & 0.798 \\
      \bottomrule
      \end{tabular}%
    }
  \end{subtable}
  \hfill
  \begin{subtable}[t]{0.545\textwidth}
    \centering
    \caption{Parameter Sensitivity Analysis to $\lambda$}
    \label{a2}
    \resizebox{\textwidth}{!}{
      \begin{tabular}{c|cccc}
      \toprule
      $\lambda$ & [0.900,1] & [0.925,1] & \textbf{[0.950,1]} & [0.975,1] \\
      \midrule
      O-AUROC & 0.759 & 0.767 & \textbf{0.786} & 0.780 \\
      P-AUROC & 0.801 & 0.784 & \textbf{0.837} & 0.828 \\
      \bottomrule
      \end{tabular}%
    }
  \end{subtable}

  \vspace{1em} 
  \begin{subtable}[t]{0.45\textwidth}
    \centering
    \caption{Parameter Sensitivity Analysis to $\sigma$}
    \label{a3}
    \resizebox{\textwidth}{!}{
      \begin{tabular}{c|cccc}
      \toprule
      $\sigma$ & [0,0.02] & [0,0.04] & \textbf{[0,0.08]} & [0,0.16] \\
      \midrule
      O-AUROC & 0.769 & 0.780 & \textbf{0.786} & 0.758 \\
      P-AUROC & 0.814 & 0.823 & \textbf{0.837} & 0.801 \\
      \bottomrule
      \end{tabular}%
    }
  \end{subtable}  
  \hfill
  \begin{subtable}[t]{0.515\textwidth}
    \centering
    \caption{Parameter Sensitivity Analysis to $\gamma$}
    \label{a4}
    \resizebox{\textwidth}{!}{
      \begin{tabular}{c|cccc}
      \toprule
      $\gamma$ & [0,0.06] & [0.03,0.09] & \textbf{[0.06,0.12]} & [0.09,0.15] \\
      \midrule
      O-AUROC & 0.711 & 0.739 & \textbf{0.786} & 0.750 \\
      P-AUROC & 0.754 & 0.788 & \textbf{0.837} & 0.799 \\
      \bottomrule
      \end{tabular}%
    }
  \end{subtable}
\end{table*}

\subsection{Parameter Sensitivity Analyses to Patches of Pseudo Anomalies}
An appropriate size is essential for learning the features of normal structures. The size of pseudo-abnormal regions is inversely correlated with the patch number $G$. As in Table~\ref{a1}, we report the ablation results regarding $G$. When $G$ was set to 16, the anomaly region was larger, resulting in lower performance, with 1.7\% and 2.7\% lower O-AUROC and P-AUROC, respectively. The anomalous region was smaller with $G$ set to 128, leading to sub-optimal performance with 2.0\% and 3.9\% lower O-AUROC and P-AUROC, respectively. This is attributed to distorted pseudo-anomalies due to too large or too small anomalous regions, and we used $G\text{=}64$ as a parameter for the comparison experiments.
 
\subsection{Parameter Sensitivity Analyses to DA-Gen}
DA-Gen contains three key parameters: $\lambda$, $\sigma$, and $\gamma$. We set a reference value based on empirical evidence and show the results of the parameter sensitivity analyses in Tables~\ref{a2}, \ref{a3}, and \ref{a4}, respectively. A smaller $\lambda$ indicates that the direction in which the anomaly occurs is shifted towards the random direction to a greater extent. When $\lambda$ was set to a smaller [0.900,1] and larger [0.975, 1], it resulted in an average decrease of 1.62\% and 2.25\% for O-AUROC and P-AUROC, respectively. We select a moderate value $\lambda=$[0.950,1] to generate the true anomaly. The stretching parameter $\sigma$ controls the degree of narrowing of the anomalous shape. We chose $\sigma=$[0,0.08] to obtain the best performance for the anomaly detection metrics of 78.6\% and 83.7\% compared to the performance degradation of 2.25\% and 2.95\% caused by the narrower $\sigma=$[0,0.02] and the wider $\sigma=$[0,0.16]. $\gamma$ represents the maximum displacement magnitude of the center of the anomaly. The $\gamma$ set to a larger [0.09,0.15] or smaller [0,0.06] resulted in an average decrease in the anomaly detection metrics of 5.5\% and 6.05\%. We choose $\gamma$=[0.06,0.12] for optimal anomaly creation. We simply visualise the generated point cloud in Appendix~\ref{Qda}.

\subsection{Parameter Sensitivity Analyses to HQC}
\begin{wraptable}{r}{0.6\textwidth} 
\vspace{1pt} 
\centering
\caption{Parameter Sensitivity Analysis to $b$.}
\label{a5}
\begin{tabular}{c|ccccc}
\toprule
$b$     & 0.150  & 0.200  & \textbf{0.250 } & 0.300  & 0.350  \\
\midrule
O-AUROC & 0.790  & 0.783  & \textbf{0.790 } & 0.779  & 0.745  \\
P-AUROC & 0.837  & 0.835  & \textbf{0.837 } & 0.820  & 0.793  \\
FPS   & 14.7  & 15.1  & \textbf{15.7} & 16.1  & 16.9  \\
\bottomrule
\end{tabular}
\vspace{1pt}
\end{wraptable}
The $b$ represents the $b$\% of samples with the lowest object-level scores at the time of the first decision that were considered normal. The proper threshold is crucial for performing HQC correctly: too small $a$ threshold leads to insignificant improvement in inference speed, and too large a threshold leads to performance degradation. The parameter sensitivity analyses to the thresholds are displayed in Table~\ref{a5}. When $b$ is set to 0.15, the efficiency gain is insignificant. When set to 0.35, its O-AUROC and P-AUROC performance decreases by 4.5\% and 4.4\%, respectively. when a is set to 0.25, the performance and efficiency of the experiment are balanced. 

\section{More Visualisation Results}
\subsection{The Distribution of Scores}
\label{scoredis}
\begin{figure}
    \centering
    \includegraphics[width=\linewidth]{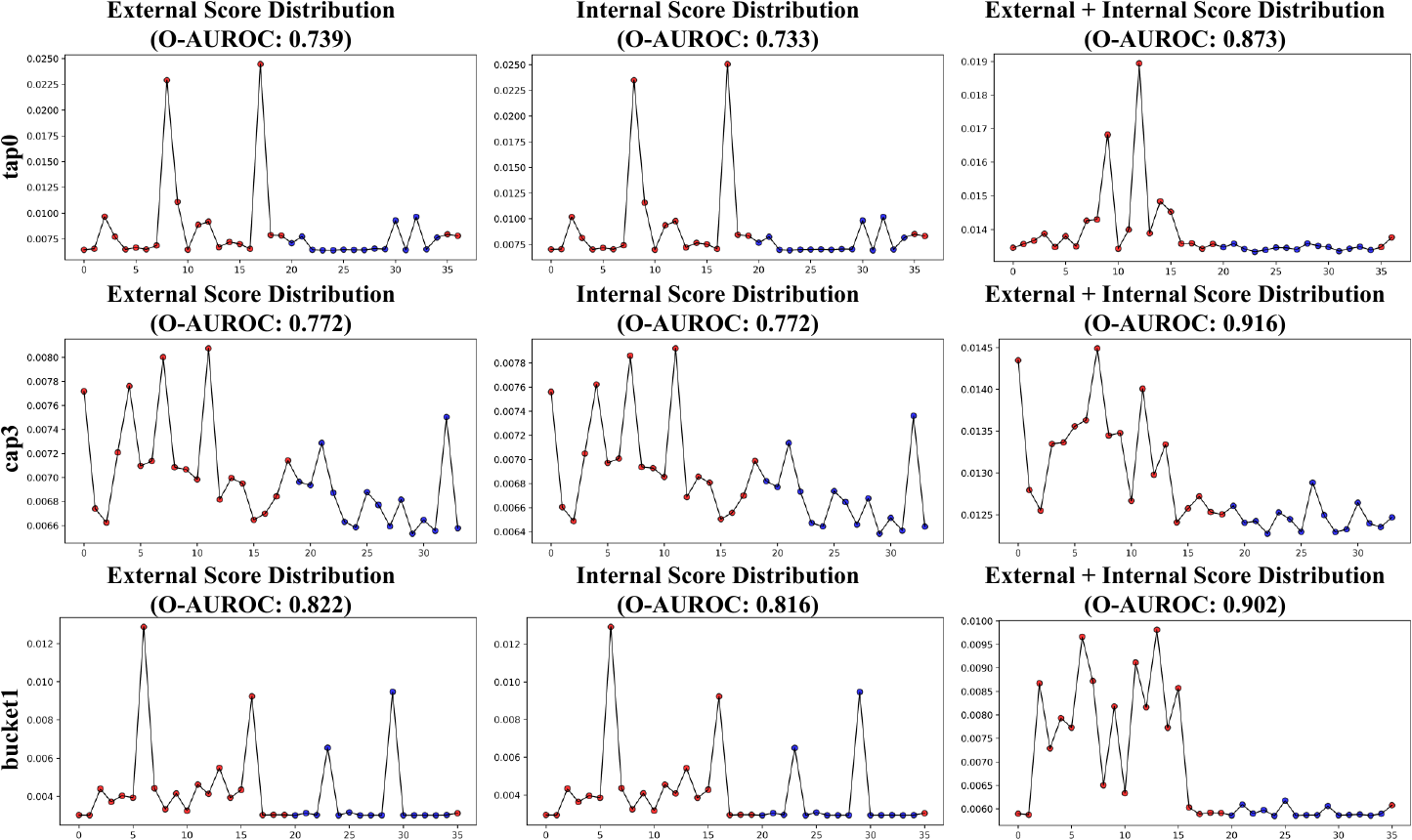}
    \caption{The distribution of anomaly detection scores on different classes, compared between External, Internal, and Internal+External.  \textcolor[rgb]{1, 0, 0}{\textbf{Red}} and \textcolor[rgb]{0, .439, .753}{\textbf{blue}} represent abnormal and normal points, respectively.}
    \label{s2}
\end{figure}

We show more distributions of Scores in Figure~\ref{s2}. The model is better able to reduce anomalies when both corrective forces are applied at the same time, e.g., the detection performance of the ``tap0'' class is 73.9\% and 73.3\% respectively, without the use of the resultant force, and the detection performance rises to 87.3\% when applied together. This is attributed to the fact that anomalies in real manufacturing processes often originate from a combination of internal and external influences.

\subsection{Qualitative Results of Detection}
\label{visdis}

\begin{figure}[!ht]
    \centering
    \includegraphics[width=0.95\linewidth]{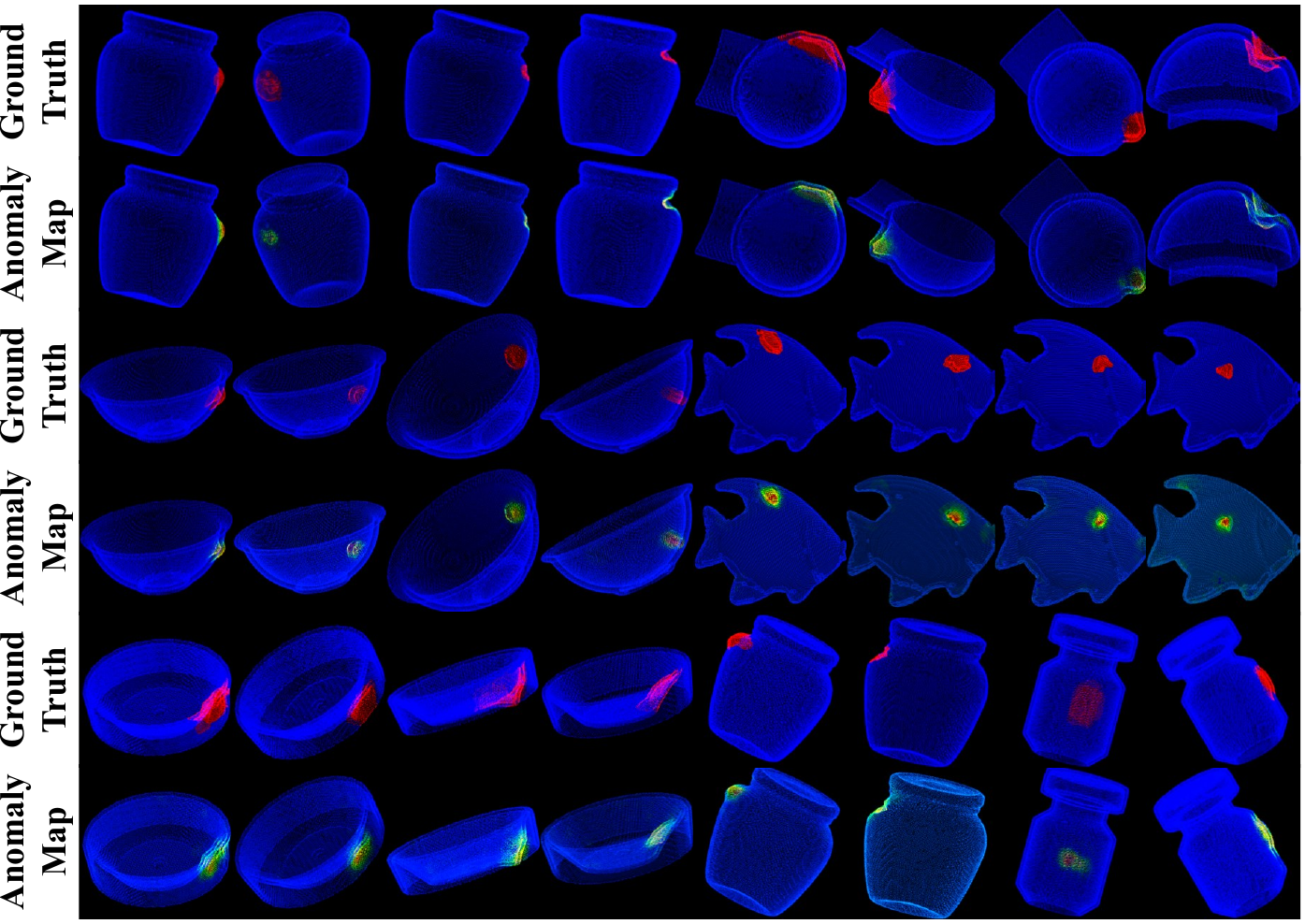}
    \caption{More qualitative results of localization.}
    \label{vis3}
\end{figure}

\begin{figure}[!ht]
    \centering
    \includegraphics[width=0.95\linewidth]{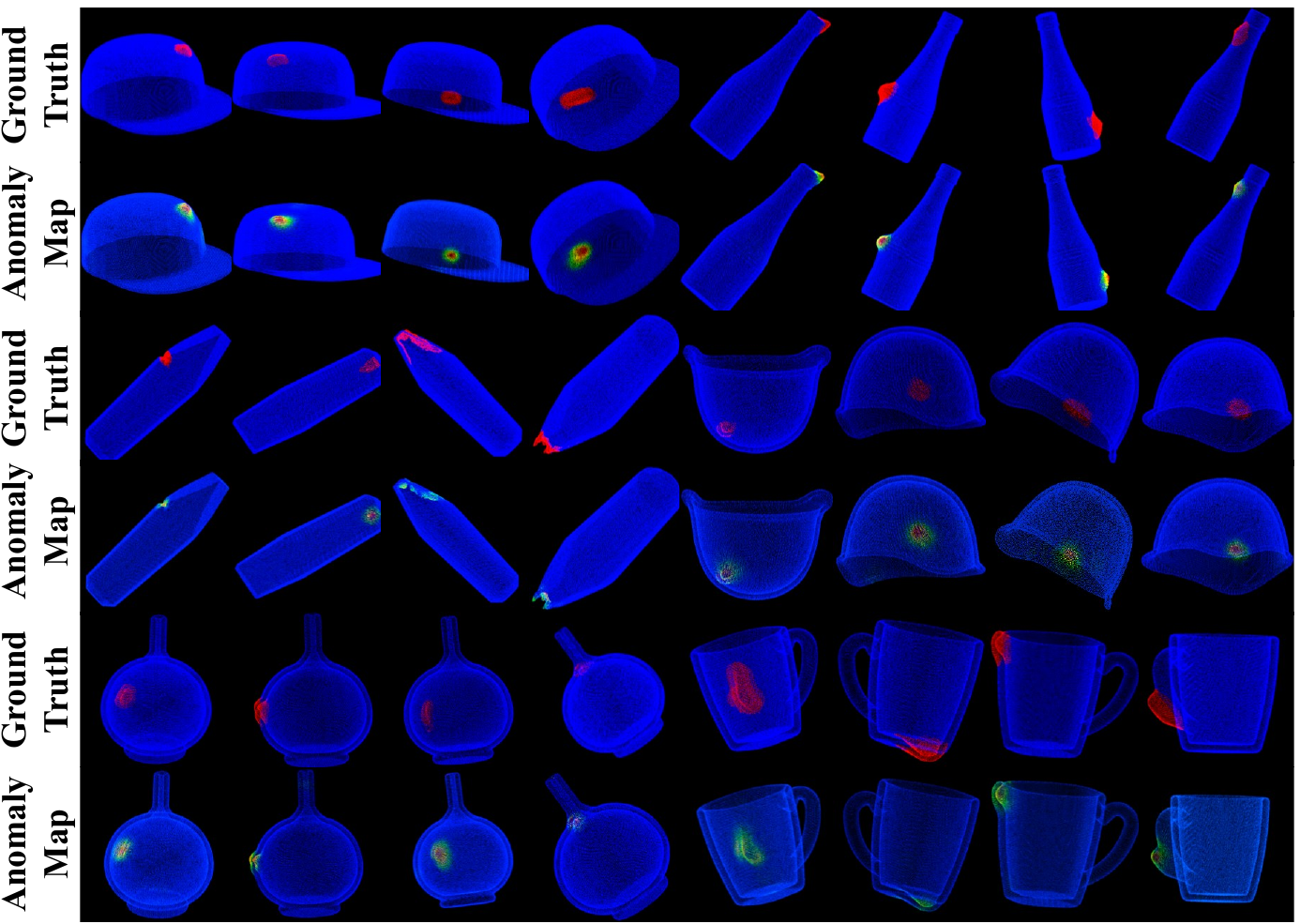}
    \caption{More qualitative results of localization.}
    \label{vis4}
\end{figure}

We show more visualisations in Figure~\ref{vis3} and \ref{vis4}. Our model successfully localises anomalies and assigns a lower anomaly score to normal structures.

\subsection{Visualisations of DA-Gen}
\label{Qda}
\begin{wrapfigure}{r}{0.5\textwidth} 
\vspace{-10pt} 
  \centering
  \includegraphics[width=\linewidth]{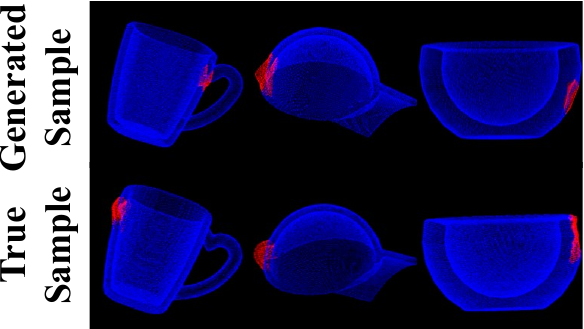}
  \caption{Qualitative results of localization.}
  \label{vis5}
\vspace{-10pt} 
\end{wrapfigure}
The realistic pseudo-anomalies facilitate our CFP-Net to capture the potential damaging forces of anomalous structures and generate corrective forces to restore the anomalies. We show the generated anomaly samples with real anomaly samples in Figure~\ref{vis5}. Our method generates anomalies close to the real ones.

\section{More Exprimental Results}
\label{moreex}
We show the P-AUROC and O-AUROC results for Anomaly-ShapeNet in Table~\ref{P-Anomaly} and Table~\ref{O-Anomaly}, respectively. We show the P-AUROC results for Anomaly-ShapeNet-New in Table~\ref{P-anomaly-new}. In Table~\ref{P-Real} we show Real3D-AD for P-AUROC. We obtained the best results in all metrics. As shown in Table~\ref{P-MVTEC}, we show Mvtec-AD for P-AUROC. We obtained the best results in all metrics.

\section{Limitations and Feature Work}
\label{future}
We discuss further limitations in this section and point the direction of future work. Our framework lacks explicit physical constraints. Current approaches in the robotics area, incorporating physical engines~\cite{barcellona2025dream} (e.g., LLM-based priors) face challenges in 3D anomaly detection, primarily due to the inherent structural ambiguity of 3D point clouds and the non-physical randomization in synthetic dataset generation. Notably, real-world benchmarks we use, like Real3D-AD~\cite{liu2023real3d} and MvTec3D-AD~\cite{Bergmann_2022} maintain authentic physical constraints. Future dataset development should integrate comprehensive physical descriptors (e.g., RGB textures, illumination conditions, and industrial contexts) to enable explicit physical regularization in anomaly detection models.

\begin{table*}[!ht]
  \centering
  \caption{O-AUROC performance of different methods on Anomaly-ShapeNet across 40 categories, where best and second-place results are highlighted in \textcolor[rgb]{1, 0, 0}{\textbf{red}} and \textcolor[rgb]{0, .439, .753}{\textbf{blue}}, respectively.}
  \label{O-Anomaly}
  \resizebox{1\textwidth}{!}{
\begin{tabular}{c|cccccccccccccc}
    \toprule
    \multicolumn{15}{c}{\textbf{O-AUROC}} \\
    \midrule
    \textbf{Method} & \textbf{ashtray0} & \textbf{bag0} & \textbf{bottle0} & \textbf{bottle1} & \textbf{bottle3} & \textbf{bowl0} & \textbf{bowl1} & \textbf{bowl2} & \textbf{bowl3} & \textbf{bowl4} & \textbf{bowl5} & \textbf{bucket0} & \textbf{bucket1} & \textbf{cap0} \\
    \midrule
    \textbf{BTF(Raw) (CVPR23’)} & 0.578  & 0.410  & 0.597  & 0.510  & 0.568  & 0.564  & 0.264  & 0.525  & 0.385  & 0.664  & 0.417  & 0.617  & 0.321  & 0.668  \\
    \textbf{BTF(FPFH) (CVPR23’)} & 0.420  & 0.546  & 0.344  & 0.546  & 0.322  & 0.509  & 0.668  & 0.510  & 0.490  & 0.609  & 0.699  & 0.401  & 0.633  & 0.618  \\
    \textbf{M3DM (CVPR23’)} & 0.577  & 0.537  & 0.574  & 0.637  & 0.541  & 0.634  & 0.663  & 0.684  & 0.617  & 0.464  & 0.409  & 0.309  & 0.501  & 0.557  \\
    \textbf{PatchCore(FPFH) (CVPR22’)} & 0.587  & 0.571  & 0.604  & 0.667  & 0.572  & 0.504  & 0.639  & 0.615  & 0.537  & 0.494  & 0.558  & 0.469  & 0.551  & 0.580  \\
    \textbf{PatchCore(PointMAE) (CVPR22')} & 0.591  & 0.601  & 0.513  & 0.601  & 0.650  & 0.523  & 0.629  & 0.458  & 0.579  & 0.501  & 0.593  & 0.593  & 0.561  & 0.589  \\
    \textbf{CPMF (PR24’)} & 0.353  & 0.643  & 0.520  & 0.482  & 0.405  & 0.783  & 0.639  & 0.625  & 0.658  & 0.683  & 0.685  & 0.482  & 0.601  & 0.601  \\
\bottomrule
    \end{tabular}%
  }
\end{table*}

\begin{table*}[!ht]
  \centering
  \caption{P-AUROC performance of different methods on Anomaly-ShapeNet across 40 categories, where best and second-place results are highlighted in \textcolor[rgb]{1, 0, 0}{\textbf{red}} and \textcolor[rgb]{0, .439, .753}{\textbf{blue}}, respectively.}
  \label{P-Anomaly}
  \resizebox{1\textwidth}{!}{
    \begin{tabular}{c|cccccccccccccc}
    \toprule
    \multicolumn{15}{c}{\textbf{P-AUROC}} \\
    \midrule
    \textbf{Method} & \textbf{ashtray0} & \textbf{bag0} & \textbf{bottle0} & \textbf{bottle1} & \textbf{bottle3} & \textbf{bowl0} & \textbf{bowl1} & \textbf{bowl2} & \textbf{bowl3} & \textbf{bowl4} & \textbf{bowl5} & \textbf{bucket0} & \textbf{bucket1} & \textbf{cap0} \\
    \midrule
    \textbf{BTF(Raw)(CVPR23’)} & 0.512  & 0.430  & 0.551  & 0.491  & 0.720  & 0.524  & 0.464  & 0.426  & 0.685  & 0.563  & 0.517  & 0.617  & 0.686  & 0.524  \\
    \textbf{BTF(FPFH)(CVPR23’)} & 0.624  & 0.746  & 0.641  & 0.549  & 0.622  & 0.710  & 0.768  & 0.518  & 0.590  & 0.679  & 0.699  & 0.401  & 0.633  & 0.730  \\
    \textbf{M3DM(CVPR23’)} & 0.577  & 0.637  & 0.663  & 0.637  & 0.532  & 0.658  & 0.663  & 0.694  & 0.657  & 0.624  & 0.489  & 0.698  & 0.699  & 0.531  \\
    \textbf{PatchCore(FPFH)(CVPR22’)} & 0.597  & 0.574  & 0.654  & 0.687  & 0.512  & 0.524  & 0.531  & 0.625  & 0.327  & 0.720  & 0.358  & 0.459  & 0.571  & 0.472  \\
    \textbf{PatchCore(PointMAE)(CVPR22’)} & 0.495  & 0.674  & 0.553  & 0.606  & 0.653  & 0.527  & 0.524  & 0.515  & 0.581  & 0.501  & 0.562  & 0.586  & 0.574  & 0.544  \\
    \textbf{CPMF(PR24’)} & 0.615  & 0.655  & 0.521  & 0.571  & 0.435  & 0.745  & 0.488  & 0.635  & 0.641  & 0.683  & 0.684  & 0.486  & 0.601  & 0.601  \\
\bottomrule
    \end{tabular}
  }
\end{table*}

\label{app1}
\begin{table*}[!ht]
  \centering
  \caption{O-AUPR performance of different methods on Anomaly-ShapeNet across 40 categories, where best and second-place results are highlighted in \textcolor[rgb]{1, 0, 0}{\textbf{red}} and \textcolor[rgb]{0, .439, .753}{\textbf{blue}}, respectively.}
  \label{Or-Anomaly}
  \resizebox{1\textwidth}{!}{
    \begin{tabular}{ccccccccccccccc}
    \toprule
    \multicolumn{15}{c}{\textbf{O-AUPR}} \\
    \midrule
    \multicolumn{1}{c|}{\textbf{Method}} & \textbf{ashtray0} & \textbf{bag0} & \textbf{bottle0} & \textbf{bottle1} & \textbf{bottle3} & \textbf{bowl0} & \textbf{bowl1} & \textbf{bowl2} & \textbf{bowl3} & \textbf{bowl4} & \textbf{bowl5} & \textbf{bucket0} & \textbf{bucket1} & \textbf{cap0} \\
    \midrule
    \multicolumn{1}{c|}{\textbf{BTF(Raw)(CVPR23’)}} & 0.578  & 0.458  & 0.466  & 0.573  & 0.543  & 0.588  & 0.464  & 0.576  & 0.654  & 0.601  & 0.615  & 0.652  & 0.620  & 0.659  \\
    \multicolumn{1}{c|}{\textbf{BTF(FPFH)(CVPR23’)}} & 0.651  & 0.551  & 0.644  & 0.625  & 0.602  & 0.576  & 0.648  & 0.515  & 0.499  & 0.632  & 0.699  & 0.483  & 0.648  & 0.618  \\
    \multicolumn{1}{c|}{\textbf{M3DM(CVPR23’)}} & 0.632  & 0.642  & 0.763  & 0.674  & 0.451  & 0.525  & 0.515  & 0.630  & 0.635  & 0.571  & 0.601  & 0.609  & 0.507  & 0.564  \\
    \multicolumn{1}{c|}{\textbf{PatchCore(FPFH)(CVPR22’)}} & 0.445  & 0.608  & 0.615  & 0.677  & 0.579  & 0.548  & 0.545  & 0.611  & 0.620  & 0.575  & 0.541  & 0.604  & 0.565  & 0.585  \\
    \multicolumn{1}{c|}{\textbf{PatchCore(PointMAE)(CVPR22’)}} & 0.679  & 0.601  & 0.545  & 0.645  & 0.651  & 0.562  & 0.611  & 0.456  & 0.556  & 0.601  & 0.585  & 0.541  & 0.642  & 0.561  \\
    \multicolumn{1}{c|}{\textbf{CPMF(PR24’)}} & 0.453  & 0.655  & 0.588  & 0.592  & 0.505  & 0.775  & 0.621  & 0.601  & 0.418  & 0.683  & 0.685  & 0.662  & 0.501  & 0.601  \\
\end{tabular}
  }
\end{table*}

\begin{table*}[!ht]
  \centering
  \caption{P-AUROC performance of different methods on Anomaly-ShapeNet-New across 12 categories, where best and second-place results are highlighted in \textcolor[rgb]{1, 0, 0}{\textbf{red}} and \textcolor[rgb]{0, .439, .753}{\textbf{blue}}, respectively.}
  \label{P-anomaly-new}
  \resizebox{1\textwidth}{!}{
    \begin{tabular}{c|cccccccccccccc}
     \toprule
    \multicolumn{14}{c}{\textbf{P-AUROC}}                                                            &  \\
    \midrule
    \textbf{Method} & \textbf{cabinet0} & \textbf{cap0} & \textbf{cap2} & \textbf{chair0} & \textbf{cup2} & \textbf{desk0} & \textbf{knife0} & \textbf{knife1} & \textbf{micro.1} & \textbf{screen0} & \textbf{vase10} & \textbf{vase6} & \textbf{Average} & \textbf{Mean Rank} \\
    \midrule
    \textbf{BTF(Raw) (CVPR23')} & 0.464  & 0.640  & 0.600  & 0.508  & 0.492  & 0.611  & 0.486  & 0.433  & 0.505  & 0.548  & 0.439  & 0.453  & 0.515  & 6.667  \\
    \textbf{BTF(FPFH) (CVPR23')} & 0.725  & 0.915  & 0.722  & 0.653  & 0.843  & 0.864  & 0.736  & 0.774  & 0.888  & 0.720  & 0.741  & 0.896  & 0.790  & 3.250  \\
    \textbf{M3DM (CVPR23')} & 0.642  & 0.754  & 0.750  & 0.532  & 0.657  & 0.587  & 0.507  & 0.601  & 0.768  & 0.613  & 0.696  & 0.668  & 0.648  & 4.750  \\
    \textbf{PatchCore(FPFH) (CVPR22')} & \textcolor[rgb]{ 0,  .439,  .753}{\textbf{0.979 }} & \textcolor[rgb]{ 0,  .439,  .753}{\textbf{0.966 }} & 0.725  & \textcolor[rgb]{ 0,  .439,  .753}{\textbf{0.852 }} & \textcolor[rgb]{ 1,  0,  0}{\textbf{0.959 }} & \textcolor[rgb]{ 0,  .439,  .753}{\textbf{0.979 }} & \textcolor[rgb]{ 0,  .439,  .753}{\textbf{0.855 }} & \textcolor[rgb]{ 1,  0,  0}{\textbf{0.935 }} & \textcolor[rgb]{ 1,  0,  0}{\textbf{0.992 }} & \textcolor[rgb]{ 0,  .439,  .753}{\textbf{0.939 }} & \textcolor[rgb]{ 1,  0,  0}{\textbf{0.948 }} & \textcolor[rgb]{ 0,  .439,  .753}{\textbf{0.950 }} & \textcolor[rgb]{ 0,  .439,  .753}{\textbf{0.923 }} & \textcolor[rgb]{ 0,  .439,  .753}{\textbf{1.917 }} \\
    \textbf{PatchCore(PointMAE) (CVPR22')} & 0.675  & 0.755  & \textcolor[rgb]{ 0,  .439,  .753}{\textbf{0.757 }} & 0.570  & 0.686  & 0.590  & 0.499  & 0.546  & 0.717  & 0.590  & 0.672  & 0.681  & 0.645  & 4.667  \\
\bottomrule
    \end{tabular}%
  }
\end{table*}

\begin{table*}[t!]
  \centering
  \caption{O-AUROC performance of different methods on Anomaly-ShapeNet-New across 12 categories, where best and second-place results are highlighted in \textcolor[rgb]{1, 0, 0}{\textbf{red}} and \textcolor[rgb]{0, .439, .753}{\textbf{blue}}, respectively.}
  \label{O-anomaly-new}
  \resizebox{1\textwidth}{!}{
    \begin{tabular}{c|cccccccccccccc}
    \toprule
    \multicolumn{14}{c}{\textbf{O-AUROC}}                                                        &  \\
    \midrule
    \textbf{Method} & \textbf{cabinet0} & \textbf{cap0} & \textbf{cap2} & \textbf{chair0} & \textbf{cup2} & \textbf{desk0} & \textbf{knife0} & \textbf{knife1} & \textbf{micro.1} & \textbf{screen0} & \textbf{vase10} & \textbf{vase6} & \textbf{Average} & \textbf{Mean Rank} \\
    \midrule
    \textbf{BTF(Raw) (CVPR23')} & 0.490  & 0.544  & 0.330  & 0.357  & 0.562  & 0.500  & 0.393  & 0.503  & 0.492  & 0.533  & 0.620  & 0.670  & 0.500  & 5.917  \\
    \textbf{BTF(FPFH) (CVPR23')} & 0.773  & 0.761  & 0.596  & 0.533  & 0.486  & 0.503  & 0.550  & 0.657  & 0.837  & 0.419  & 0.788  & 0.676  & 0.632  & 4.250  \\
    \textbf{M3DM (CVPR23')} & 0.613  & 0.625  & 0.435  & 0.543  & 0.667  & 0.580  & 0.487  & 0.507  & 0.483  & 0.663  & 0.612  & 0.667  & 0.574  & 4.583  \\
    \textbf{PatchCore(FPFH) (CVPR22')} & \textcolor[rgb]{ 1,  0,  0}{\textbf{1.000 }} & \textcolor[rgb]{ 1,  0,  0}{\textbf{0.944 }} & \textcolor[rgb]{ 1,  0,  0}{\textbf{0.870 }} & \textcolor[rgb]{ 1,  0,  0}{\textbf{0.903 }} & \textcolor[rgb]{ 1,  0,  0}{\textbf{0.962 }} & \textcolor[rgb]{ 0,  .439,  .753}{\textbf{0.967 }} & \textcolor[rgb]{ 0,  .439,  .753}{\textbf{0.863 }} & \textcolor[rgb]{ 0,  .439,  .753}{\textbf{0.720 }} & \textcolor[rgb]{ 1,  0,  0}{\textbf{0.917 }} & \textcolor[rgb]{ 0,  .439,  .753}{\textbf{0.667 }} & \textcolor[rgb]{ 0,  .439,  .753}{\textbf{0.948 }} & \textcolor[rgb]{ 0,  .439,  .753}{\textbf{0.848 }} & \textcolor[rgb]{ 0,  .439,  .753}{\textbf{0.884 }} & \textcolor[rgb]{ 1,  0,  0}{\textbf{1.500 }} \\
    \textbf{PatchCore(PointMAE) (CVPR22')} & 0.440  & 0.667  & 0.565  & 0.577  & 0.619  & 0.543  & 0.490  & 0.490  & 0.517  & 0.656  & 0.617  & 0.709  & 0.574  & \textcolor[rgb]{ 0,  .439,  .753}{\textbf{4.333 }} \\
\bottomrule
    \end{tabular}%
  }
\end{table*}

\begin{table*}[t!]
  \centering
  \caption{O-AUROC performance of different methods on only-3D MvTec3D-AD across 10 categories, where best and second-place results are highlighted in \textcolor[rgb]{1, 0, 0}{\textbf{red}} and \textcolor[rgb]{0, .439, .753}{\textbf{blue}}, respectively.}
  \label{O-MVTEC}
  \resizebox{1\textwidth}{!}{
    \begin{tabular}{c|cccccccccccc}
    \toprule
    \multicolumn{12}{c}{\textbf{O-AUROC}}                                                         &  \\
    \midrule
    \textbf{Method} & \textbf{Bagel} & \textbf{Cable Gland} & \textbf{Carrot} & \textbf{Cookie} & \textbf{Dowel} & \textbf{Foam} & \textbf{Peach} & \textbf{Potato} & \textbf{Rope} & \textbf{Tire} & \textbf{Average} & \textbf{Mean Rank} \\
    \midrule
    \textbf{BTF (CVPR23')} & 0.825  & 0.551  & 0.952  & 0.797  & 0.883  & 0.582  & 0.758  & 0.889  & 0.929  & 0.653  & 0.782  & 6.400  \\
    \textbf{EasyNet (MM23')} & 0.629  & 0.716  & 0.768  & 0.731  & 0.660  & 0.710  & 0.712  & 0.711  & 0.688  & 0.731  & 0.706  & 7.100  \\
    \textbf{M3DM (CVPR23')} & 0.941  & 0.651  & 0.965  & 0.969  & 0.905  & 0.760  & 0.880  & 0.974  & 0.926  & 0.765  & 0.874  & 4.400  \\
    \textbf{AST (WACV23')} & 0.881  & 0.576  & 0.965  & 0.957  & 0.679  & 0.797  & \textcolor[rgb]{ 0,  .439,  .753}{\textbf{0.990 }} & 0.915  & 0.956  & 0.611  & 0.833  & 4.900  \\
    \textbf{Shape-Guided (ICML23')} & \textcolor[rgb]{ 0,  .439,  .753}{\textbf{0.983 }} & 0.682  & \textcolor[rgb]{ 0,  .439,  .753}{\textbf{0.978 }} & \textcolor[rgb]{ 1,  0,  0}{\textbf{0.998 }} & \textcolor[rgb]{ 0,  .439,  .753}{\textbf{0.960 }} & 0.737  & \textcolor[rgb]{ 1,  0,  0}{\textbf{0.993 }} & \textcolor[rgb]{ 0,  .439,  .753}{\textbf{0.979 }} & 0.966  & 0.871  & 0.915  & 2.600  \\
    \textbf{CFM (CVPR24')} & \textbackslash{} & \textbackslash{} & \textbackslash{} & \textbackslash{} & \textbackslash{} & \textbackslash{} & \textbackslash{} & \textbackslash{} & \textbackslash{} & \textbackslash{} & 0.816  & \textbackslash{} \\
\bottomrule
    \end{tabular}%
}
\end{table*}

\begin{table*}[th!]
  \centering
  \caption{P-AUROC performance of different only-3D methods on MvTec3D-AD across 10 categories, where best and second-place results are highlighted in \textcolor[rgb]{1, 0, 0}{\textbf{red}} and \textcolor[rgb]{0, .439, .753}{\textbf{blue}}, respectively.}
  \label{P-MVTEC}
  \resizebox{1\textwidth}{!}{
    \begin{tabular}{c|cccccccccccc}
    \toprule
    \multicolumn{12}{c}{\textbf{P-AUROC}}                                                         &  \\
    \midrule
    \textbf{Method} & \textbf{Bagel} & \textbf{Cable Gland} & \textbf{Carrot} & \textbf{Cookie} & \textbf{Dowel} & \textbf{Foam} & \textbf{Peach} & \textbf{Potato} & \textbf{Rope} & \textbf{Tire} & \textbf{Average} & \textbf{Mean Rank} \\
    \midrule
    \textbf{BTF (CVPR23')} & 0.973  & 0.879  & 0.982  & 0.906  & 0.892  & 0.735  & 0.977  & 0.982  & 0.956  & 0.961  & 0.924  & 4.800  \\
    \textbf{M3DM (CVPR23')} & 0.943  & 0.818  & 0.977  & 0.882  & 0.881  & 0.743  & 0.958  & 0.974  & 0.950  & 0.929  & 0.906  & 6.900  \\
    \textbf{Shape-Guided (ICML23')} & \textcolor[rgb]{ 0,  .439,  .753}{\textbf{0.974 }} & 0.871  & 0.981  & \textcolor[rgb]{ 0,  .439,  .753}{\textbf{0.924 }} & 0.898  & \textcolor[rgb]{ 0,  .439,  .753}{\textbf{0.773 }} & 0.978  & \textcolor[rgb]{ 0,  .439,  .753}{\textbf{0.983 }} & 0.955  & 0.969  & 0.931  & 3.800  \\
    \textbf{CPMF (PR24')} & 0.958  & \textcolor[rgb]{ 1,  0,  0}{\textbf{0.946 }} & 0.979  & 0.868  & 0.897  & 0.746  & 0.980  & 0.981  & 0.961  & 0.977  & 0.929  & 4.300  \\
    \textbf{MMRD (AAAI24')} & 0.926  & 0.806  & 0.965  & 0.858  & 0.904  & 0.731  & 0.962  & 0.958  & 0.966  & 0.936  & 0.901  & 6.600  \\
    \textbf{3DSR (WACV24')} & 0.922  & 0.872  & \textcolor[rgb]{ 1,  0,  0}{\textbf{0.984 }} & 0.859  & \textcolor[rgb]{ 1,  0,  0}{\textbf{0.940 }} & 0.714  & 0.970  & 0.978  & \textcolor[rgb]{ 0,  .439,  .753}{\textbf{0.977 }} & 0.958  & 0.907  & 5.000  \\
\bottomrule
    \end{tabular}%
}
\end{table*}
\begin{table*}[!ht]
  \centering
  \caption{P-AUROC performance of different methods on Real3D-AD across 12 categories, where best and second-place results are highlighted in \textcolor[rgb]{1, 0, 0}{\textbf{red}} and \textcolor[rgb]{0, .439, .753}{\textbf{blue}}, respectively.}
  \label{P-Real}
  \resizebox{1\textwidth}{!}{
     \begin{tabular}{c|cccccccccccccc}
    \toprule
    \multicolumn{15}{c}{\textbf{P-AUROC}} \\
    \midrule
    \textbf{Method} & \textbf{Airplane} & \textbf{Car} & \textbf{Candy} & \textbf{Chicken} & \textbf{Diamond} & \textbf{Duck} & \textbf{Fish} & \textbf{Gemstone} & \textbf{Seahorse} & \textbf{Shell} & \textbf{Starfish} & \textbf{Toffees} & \textbf{Mean} & \textbf{Mean Rank} \\
    \midrule
    \textbf{BTF(Raw) (CVPR23')} & 0.564  & 0.647  & 0.735  & 0.609  & 0.563  & 0.601  & 0.514  & 0.597  & 0.520  & 0.489  & 0.392  & 0.623  & 0.571  & 7.417  \\
    \textbf{BTF(FPFH) (CVPR23')} & 0.738  & 0.708  & \textcolor[rgb]{ 0,  .439,  .753}{\textbf{0.864 }} & 0.735  & \textcolor[rgb]{ 0,  .439,  .753}{\textbf{0.882 }} & \textcolor[rgb]{ 0,  .439,  .753}{\textbf{0.875 }} & 0.709  & 0.891  & 0.512  & 0.571  & 0.501  & \textcolor[rgb]{ 0,  .439,  .753}{\textbf{0.815 }} & 0.733  & 4.417  \\
    \textbf{M3DM (CVPR23')} & 0.547  & 0.602  & 0.679  & 0.678  & 0.608  & 0.667  & 0.606  & 0.674  & 0.560  & 0.738  & 0.532  & 0.682  & 0.631  & 6.417  \\
    \textbf{PatchCore(FPFH) (CVPR22')} & 0.562  & 0.754  & 0.780  & 0.429  & 0.828  & 0.264  & 0.829  & \textcolor[rgb]{ 1,  0,  0}{\textbf{0.910 }} & 0.739  & 0.739  & 0.606  & 0.747  & 0.682  & 5.000  \\
    \textbf{PatchCore(PointMAE) (CVPR22')} & 0.569  & 0.609  & 0.627  & 0.729  & 0.718  & 0.528  & 0.717  & 0.444  & 0.633  & 0.709  & 0.580  & 0.580  & 0.620  & 6.417  \\
    \textbf{Reg3D-AD (NeurIPS23')} & 0.631  & 0.718  & 0.724  & 0.676  & 0.835  & 0.503  & 0.826  & 0.545  & \textcolor[rgb]{ 0,  .439,  .753}{\textbf{0.817 }} & 0.811  & 0.617  & 0.759  & 0.705  & 4.667  \\
\bottomrule
    \end{tabular}%
  }
\end{table*}

\begin{table*}[!th]
  \centering
  \caption{O-AUROC and P-AUROC performances of different methods on Anomaly-IntraVariance across 16 categories, where best and second-place results are highlighted in \textcolor[rgb]{1, 0, 0}{\textbf{red}} and \textcolor[rgb]{0, .439, .753}{\textbf{blue}}, respectively.}
  \label{intra}
  \resizebox{\textwidth}{!}{
      \begin{tabular}{c|ccccccccc|ccccccccc}
    \toprule
    \multicolumn{19}{c}{\textbf{Group 1 (two subcategories)}} \\
    \midrule
    \multirow{2}[4]{*}{\textbf{Method}} & \multicolumn{9}{c|}{\textbf{O-AUROC}}                                 & \multicolumn{9}{c}{\textbf{P-AUROC}} \\
\cmidrule{2-19}          & \textbf{bucket} & \textbf{micro} & \textbf{cup} & \textbf{tap} & \textbf{bottle} & \textbf{helmet} & \textbf{cap} & \textbf{bowl} & \textbf{Average} & \textbf{bucket} & \textbf{micro} & \textbf{cup} & \textbf{tap} & \textbf{bottle} & \textbf{helmet} & \textbf{cap} & \textbf{bowl} & \textbf{Average} \\
    \midrule
    \textbf{BTF (Raw)} & 0.486  & 0.524  & 0.530  & 0.493  & 0.522  & 0.592  & 0.439  & 0.544  & 0.516  & 0.498  & 0.553  & 0.520  & 0.486  & 0.571  & 0.520  & 0.458  & 0.465  & 0.509  \\
    \textbf{BTF (FPFH)} & 0.473  & 0.723  & 0.500  & 0.587  & 0.680  & 0.433  & 0.661  & 0.700  & 0.595  & 0.730  & 0.831  & 0.811  & \textcolor[rgb]{ 0,  .439,  .753}{\textbf{0.606 }} & 0.684  & 0.702  & 0.936  & 0.786  & 0.761  \\
    \textbf{M3DM (PointMAE)} & 0.555  & 0.592  & 0.506  & 0.436  & 0.624  & 0.500  & 0.533  & 0.563  & 0.539  & 0.537  & 0.714  & 0.588  & 0.494  & 0.663  & 0.561  & 0.662  & 0.589  & 0.601  \\
    \textbf{PatchCore (FPFH)} & 0.688  & \textcolor[rgb]{ 1,  0,  0}{\textbf{0.838 }} & 0.692  & 0.533  & \textcolor[rgb]{ 0,  .439,  .753}{\textbf{0.808 }} & 0.588  & \textcolor[rgb]{ 1,  0,  0}{\textbf{0.941 }} & \textcolor[rgb]{ 1,  0,  0}{\textbf{0.807 }} & 0.737  & \textcolor[rgb]{ 0,  .439,  .753}{\textbf{0.803 }} & \textcolor[rgb]{ 1,  0,  0}{\textbf{0.869 }} & \textcolor[rgb]{ 0,  .439,  .753}{\textbf{0.830 }} & 0.594  & 0.668  & \textcolor[rgb]{ 1,  0,  0}{\textbf{0.743 }} & \textcolor[rgb]{ 1,  0,  0}{\textbf{0.971 }} & \textcolor[rgb]{ 0,  .439,  .753}{\textbf{0.919 }} & \textcolor[rgb]{ 0,  .439,  .753}{\textbf{0.800 }} \\
\bottomrule
    \end{tabular}%
    }
\end{table*}

\clearpage


\bibliographystyle{IEEEtran}
\end{document}